\documentclass[pdflatex,sn-mathphys-num]{sn-jnl}% Math and Physical Sciences Numbered Reference Style
\usepackage{graphicx}%
\usepackage{multirow}%
\usepackage{amsmath,amssymb,amsfonts}%
\usepackage{amsthm}%
\usepackage{mathrsfs}%
\usepackage[title]{appendix}%
\usepackage{xcolor}%
\usepackage{textcomp}%
\usepackage{manyfoot}%
\usepackage{booktabs}%
\usepackage{algorithm}%
\usepackage{algorithmicx}%
\usepackage{algpseudocode}%
\usepackage{listings}%
\usepackage{graphicx}
\usepackage{comment}
\usepackage{siunitx}
\usepackage{amsmath}
\theoremstyle{thmstyleone}%
\theoremstyle{thmstyletwo}%

\theoremstyle{thmstylethree}%

\begin{document}

\title[Article Title]{Reconfiguration of supernumerary robotic limbs for human augmentation}


\author[1]{\fnm{Mustafa} \sur{Mete}}\email{a.mustafamete@gmail.com}
\equalcont{These authors contributed equally to this work.}

\author[1,2]{\fnm{Anastasia} \sur{Bolotnikova}}\email{anastasia.bolotnikova@laas.fr}
\equalcont{These authors contributed equally to this work.}

\author[1]{\fnm{Alexander} \sur{Schüßler}}\email{alexander.schuessler@epfl.ch}
\equalcont{These authors contributed equally to this work.}

\author*[1]{\fnm{Jamie} \sur{Paik}}\email{jamie.paik@epfl.ch}

\affil[1]{\orgdiv{Reconfigurable Robotics Lab}, \orgname{École Polytechnique Fédérale de Lausanne (EPFL)}, \orgaddress{\city{Lausanne}, \country{Switzerland}}}

\affil[2]{\orgdiv{Laboratory for Analysis and Architecture of Systems (LAAS)}, \orgname{CNRS}, \orgaddress{\city{Toulouse}, \country{France}}}


\abstract{Wearable robots aim to seamlessly adapt to humans and their environment with personalized interactions. Existing supernumerary robotic limbs (SRLs), which enhance the physical capabilities of humans with additional extremities, have thus far been developed primarily for task-specific applications in structured industrial settings, limiting their adaptability to dynamic and unstructured environments. Here, we introduce a novel reconfigurable SRL framework grounded in a quantitative analysis of human augmentation to guide the development of more adaptable SRLs for diverse scenarios. This framework captures how SRL configuration shapes workspace extension and human–robot collaboration. We define human augmentation ratios to evaluate collaborative, visible extended, and non-visible extended workspaces, enabling systematic selection of SRL placement, morphology, and autonomy for a given task. Using these metrics, we demonstrate how quantitative augmentation analysis can guide the reconfiguration and control of SRLs to better match task requirements. We validate the proposed approach through experiments with a reconfigurable SRL composed of origami-inspired modular elements. Our results suggest that reconfigurable SRLs, informed by quantitative human augmentation analysis, offer a new perspective for providing adaptable human augmentation and assistance in everyday environments.}


\maketitle

\section{Introduction}
\begin{figure}[]
    \centering
    \includegraphics[width=1\linewidth]{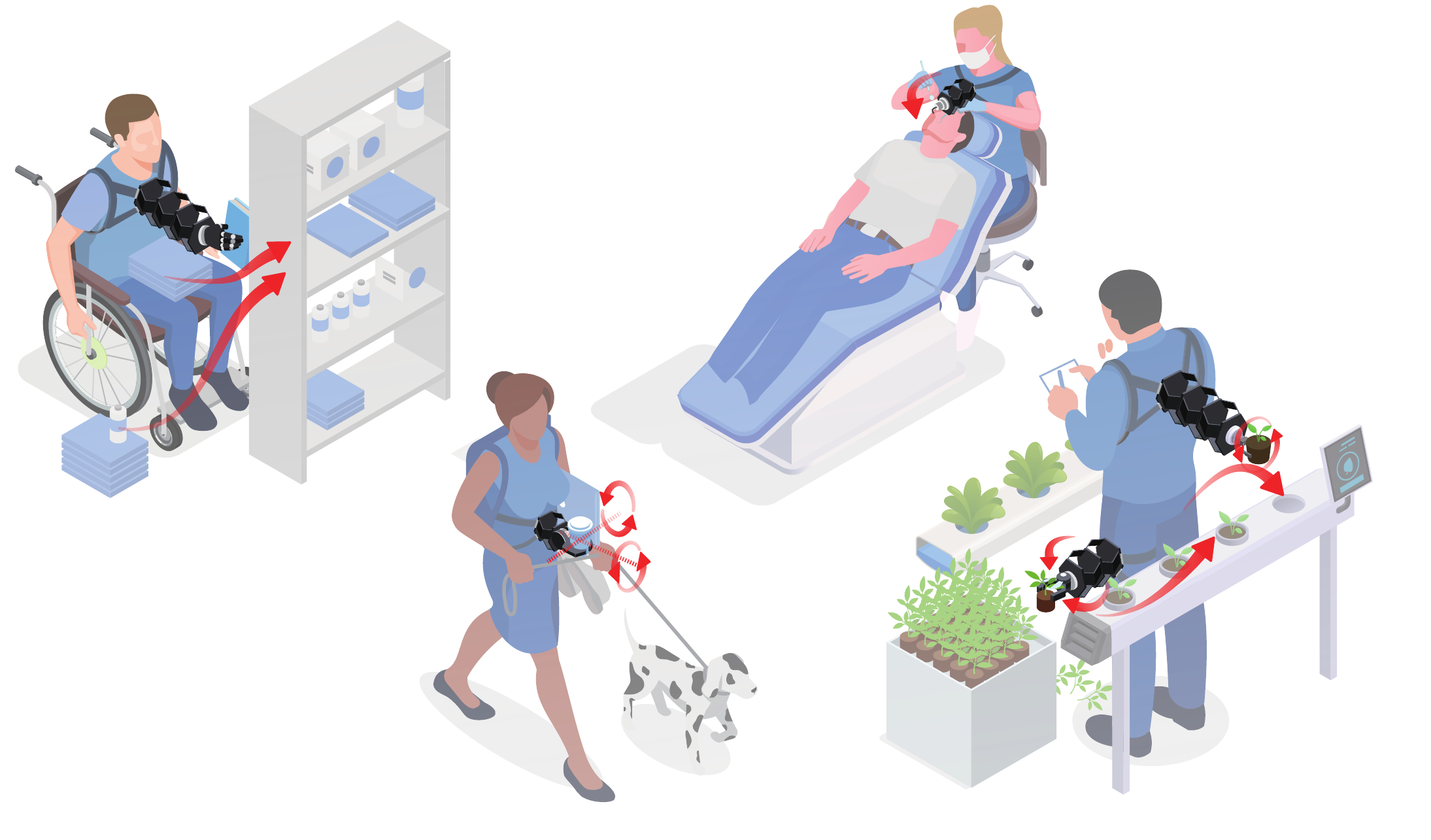}
    \caption{\textbf{Human augmentation with reconfigurable supernumerary robotic limbs. }
    Human augmentation through supernumerary robotic limbs (SRLs) has the potential to provide personalized assistance across diverse scenarios. SRLs enhance human manipulation capabilities by enabling tasks such as object stabilization when the user’s arms are occupied. They support dentists and surgeons by facilitating multi-arm procedures and improving productivity in industrial contexts, such as crop planting. Reconfigurable SRLs adapt their morphology, placement, and level of autonomy to the task and scenario, providing assistance that ranges from manual to fully autonomous task execution.}
    \label{fig:1:vision}
\end{figure}

% I1 - Human Augmentation 
Human augmentation with robotics focuses on extending the physical capabilities of humans beyond natural limitations for applications such as assistive robotics \cite{bergamasco2016, gao2025, ferroni2025, tricomi2024}, surgery \cite{goldberg2024,abdi2016}, manufacturing \cite{parietti2016, Parietti2014_1, Parietti2014_2}, and space exploration \cite{Ballesteros2023, ballesteros2024} (Fig. \ref{fig:1:vision}).
% kennedy2023

% I2 - Supernumerary Robotic Limbs for Human Augmenation
\textbf{Supernumerary robotic limbs (SRLs)} are wearable robotic devices that extend human physical capabilities by adding artificial limbs to the human body \cite{prattichizzo2021,yang2021,eden2022}. These SRLs include additional fingers \cite{Kieliba2021,clode2024,franco2021,cunningham2018,hussain2016,hussain2017}, arms \cite{Parietti2014_1,Parietti2014_2,dominijanni2023human,Ballesteros2023,veronneau2020,saraiji2018_metarms,sasaki2017_metalimbs,llorensbonilla2012,zhang2022,yamamura2023}, tails \cite{nabeshima2019}, and legs \cite{parietti2016,treers2017,haoSupernumeraryRoboticLimbs2020}. Yet, current SRLs are primarily limited to structured industrial or laboratory environments due to challenges in adapting to different operators and environments under dynamic and unstructured conditions.

% for stabilizing, lifting, or manipulating objects \cite{Parietti2014_1, Parietti2014_2, LlorensBonilla2014, Veronneau2019, Ballesteros2023}

% I3 - SRL with soft/origami materials 
Most existing SRLs are made mainly from rigid, bulky components, which limits their adaptability to the human body and the surrounding environment. This raises safety concerns and reduces wearability and user comfort in daily use. Researchers address this challenge by introducing soft SRLs \cite{Nguyen2019_1, Nguyen2019_2, ciullo2020, zhang2024}, offering enhanced safety thanks to their inherent material-based compliance. However, these SRLs, made of fully soft bodies and actuators, suffer from significant undesired deformations, which pose control challenges, and they can be bulky due to components such as fluidic cables, valves, and compressors. As an alternative, researchers propose folding-based soft SRLs for more compact, lightweight, and structurally reconfigurable SRL designs \cite{Robertson2021, mete2021, Kusunoki2023, Zhang2023, Liu2021, kim2025}, which improve SRL wearability and storage. However, optimal placement of these SRLs on the human body, as well as adaptation of their control to different configurations, scenarios, and tasks, remains a challenge.

% I4 - Control of SRL
Previous research includes various control interfaces for SRLs, spanning a spectrum with two extreme cases: fully manual and fully autonomous control. We define fully manual control as control of each SRL degree of freedom (DoF) through human intention for a specific task, whereas fully autonomous control refers to the SRL planning and taking actions for a task autonomously using embedded sensors, actuators, and processors. Control methods between these two extremes combine intention-detection interfaces with autonomous control.

Manual control methods can generally be grouped into two categories \cite{dominijanni2021,eden2022}: null-space control and skill-transfer control. In null-space control, users utilize DoFs that are not used for natural movement to control the SRLs. This enables true full-DoF extension of the human plus the SRL. Researchers explore neural interfaces \cite{penaloza2018,tang2022}, muscular interfaces such as the diaphragm \cite{dominijanni2023human} and auricular muscles \cite{leal2025neuromuscular}, and kinematic interfaces such as arm motion \cite{lisinibaldi2025} and fingertip forces \cite{guggenheim2020}. However, the total motor-task null space is limited. When the task requires more DoFs than available null-space interfaces, users re-use existing body DoFs that are irrelevant to the desired task to control the SRL; this is referred to as skill-transfer control. Existing skill-transfer interfaces include joysticks \cite{Nguyen2019_1}, foot interfaces \cite{amanhoud2021, Kieliba2021, saraiji2018_metarms, sasaki2017_metalimbs}, and electromyography (EMG) sensors \cite{hussain2016,maimeri2019}. Both manual control methods require sensory feedback to execute a task and impose a high cognitive load on the human operator during complex tasks.

Autonomous control strategies, on the other hand, reduce the required control DoFs and cognitive load for human operators by autonomously executing entire tasks or sub-tasks, such as object detection \cite{tang2022,luo2025} and force assistance \cite{amanhoud2021}. Although these strategies may offer significant benefits for full-DoF extension, they can fail in tasks requiring continuous knowledge of human intention and may reduce the sense of embodiment of extra limbs. Therefore, the ability to select across the spectrum of manual and autonomous control for different tasks remains necessary and is an ongoing challenge for SRL adaptability under unknown conditions.

This study presents reconfigurable SRLs that adapt to varying users and tasks by introducing a human augmentation analysis and an SRL reconfiguration and control strategy. In the augmentation analysis, we quantify the extended and collaborative manipulation workspace of natural arms and SRLs as a function of SRL morphology, placement, and natural human sensory input, such as visual feedback. This analysis informs the reconfiguration and control strategies for adaptable SRLs. In the reconfiguration and control strategies, we choose the SRL structure, spatial distribution, placement on the body, and the level of autonomy between manual and autonomous control based on different users and tasks. To validate our approach on a real system, we propose a modular, origami-inspired robotic arm that provides high mechanical reconfigurability and improved wearability with a lightweight, compact form factor. Our quantitative experimental results validate the adaptability of the proposed SRL approach across tasks and scenarios, such as object picking, manipulation, and stabilization. By introducing adaptability to SRLs, our approach offers a new perspective on human augmentation and assistance in everyday environments (Fig. \ref{fig:1:vision}, Movie S1).

\section{Results}\label{sec2}
\subsection{Human Augmentation Analysis}
% INTRO WHY?
To effectively reconfigure SRL morphology, choose SRL placement on the human body, and select an appropriate control strategy, it is essential to understand and quantify SRL interaction with the human sensorimotor system. Prior work has studied workspace analysis of SRLs and the human wearer to inform SRL design \cite{nakabayashi2017, liao2023}. Yet, analyzing limb workspaces alone is not sufficient for selecting an optimal control method. To determine whether sensorimotor control of SRLs is feasible, we must know whether sufficient sensory feedback about the SRL position and state is available to perform a given task.

We introduce a general human augmentation analysis that determines extended and collaborative workspaces while accounting for the human sensory-feedback workspace. For sensory feedback, we focus on visual feedback because it is a primary modality informing the user about the SRL position and state. Feedback could also be provided through other modalities, such as tactile feedback, or other sensing methods embedded in the SRL and conveyed to the human sensorimotor system. In such cases, the sensory-feedback workspace can be updated to incorporate the additional sensory information.

% explain workspaces
We define the following workspaces. The augmented human workspace $A$ is the union of the human arms workspace $H$ and the SRL workspace $R$. The collaborative workspace $C$, defined as the intersection of $R$ and $H$, is where collaborative tasks requiring two- or three-handed manipulation can be achieved. The visible extended workspace $E_{\text{V}}$ corresponds to the portion of the SRL workspace that extends the human workspace within the wearer’s visual field $V$, whereas the non-visible extended workspace $E_{\text{NV}}$ corresponds to the portion that extends the human workspace outside the visual field $V$. In $E_{\text{V}}$ and $C$, the wearer can receive visual feedback and can therefore achieve sensorimotor control of the SRL. In contrast, in $E_{\text{NV}}$, the wearer lacks visual feedback about the SRL and the task. Therefore, sensorimotor control is challenging and unreliable without additional feedback; thus autonomous control is typically required. This decomposition separates SRL workspace into regions where closed-loop human control is feasible versus regions better suited for autonomous behaviors.

% WHAT AUGMENTATION RATIO
For the augmentation-ratio analysis, we consider SRLs as modular units with $n$ modules and kinematics $M$. We quantify the \textbf{augmentation ratios} as follows:

%\begin{align}
\begin{equation}
\begin{split}
   r_e(n, M) &= \frac{A(n, M)}{H}, \\
   r_{e,v}(n, M) &= \frac{E_{V}(n, M) \cup H}{H}, \\
   r_c(n, M) &= \frac{C(n, M)}{H}. 
\end{split}
\end{equation}
%\end{align}

The relations between the workspaces are formalized as follows:
\begin{equation}
\begin{split}
    A &= R \cup H, \quad C = R \cap H, \quad R = C \cup E_{\text{V}} \cup E_{\text{NV}}, \\
    \quad E_{\text{V}} &= (R \setminus H) \cap V, \quad E_{\text{NV}} = R \setminus (E_{\text{V}} \cup H).
\end{split}
\label{eq:workspace_calculation}
\end{equation}

The extension ratio, $\boldsymbol{r_e}$, compares the augmented workspace $A$ with the natural human workspace $H$ and estimates the overall augmentation provided by a given SRL configuration. We also compute the visible extension ratio, $\boldsymbol{r_{e,v}}$, which includes only the visible extended workspace $E_{\text{V}}$ together with $H$. The collaboration ratio, $\boldsymbol{r_c}$, measures how much of the SRL workspace overlaps with $H$, approximating the space in which collaborative tasks between the human arms and the SRL are possible.

We also define the \textbf{workspace ratios}, $\boldsymbol{w_{i,n}}$, as a function of the number of modules $n$ (from 1 to $N$):

\begin{equation}
    w_{i,n} = \frac{i}{R} : i \in \{C, E_\text{V}, E_{\text{NV}}\}, n \in \{1,2,...,N\},
    \sum_i w_{i,n} = 1.0. \\
\end{equation}

These ratios describe the relative distribution of $C$, $E_{\text{V}}$, and $E_{\text{NV}}$ within $R$.

% SUMMARY FOR QUANTIFICATION
In summary, the human augmentation ratios and SRL workspace ratios quantify workspace extension and collaboration across SRL configurations. The proposed reconfiguration and control strategy uses these quantities to adapt SRL structure and control accordingly.

\begin{figure}[]
    \centering
    \includegraphics[width=\linewidth]{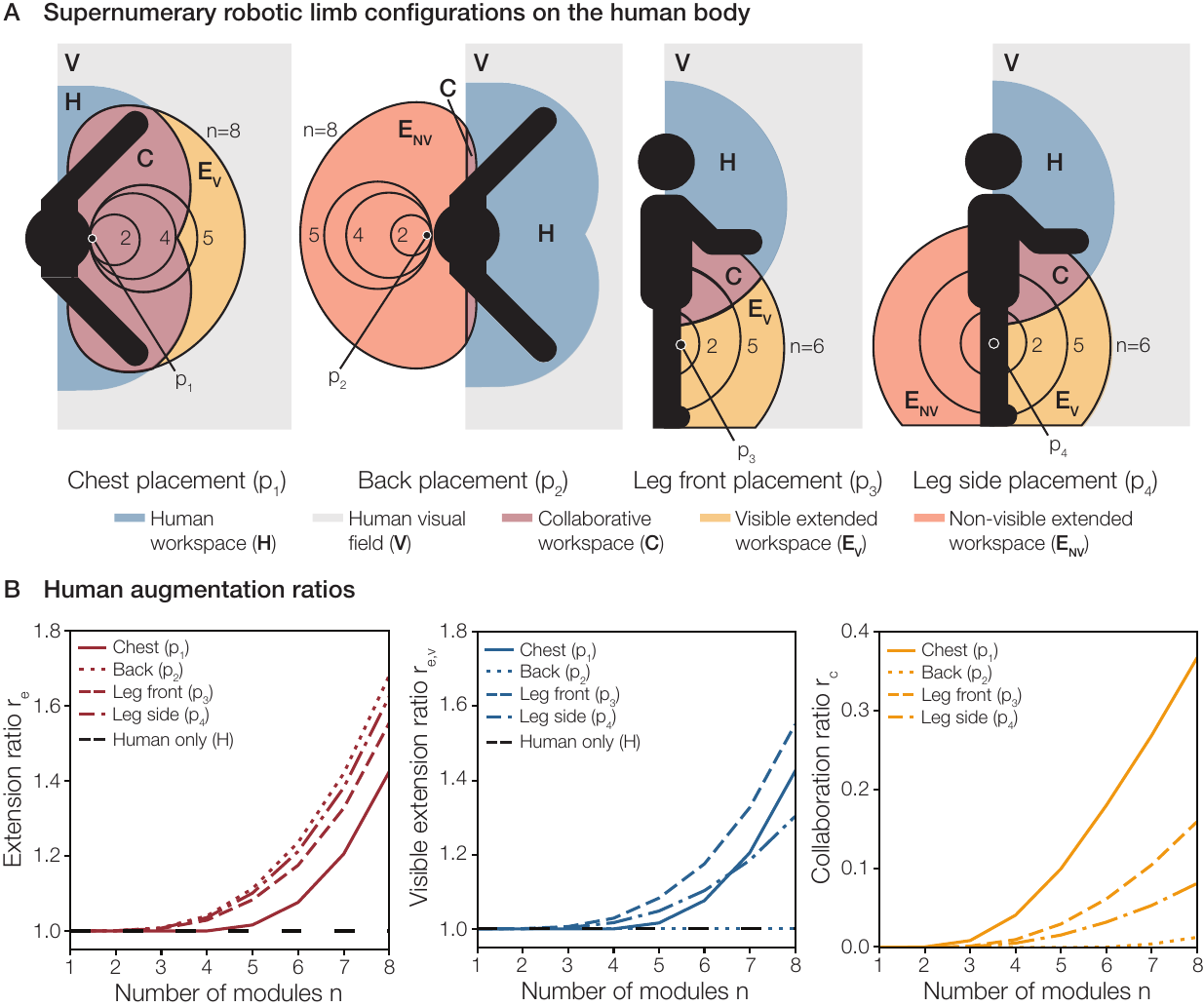}
    \caption{\textbf{Human augmentation analysis. }Different tasks and scenarios require varying configurations of SRLs. This analysis quantifies the impact of varying placement $p_j$ and morphology (number of modules $n$) of the SRL, as well as the human visual field $V$ on human augmentation. (\textbf{A}) The 4 different SRL configurations on the human body, which capture the range of potential augmentation tasks enabled by the SRL. The SRL is placed on the chest $p_1$, back $p_2$, front of the leg $p_3$, and side of the leg $p_4$. (\textbf{B}) The human augmentation ratios as a function of the number of modules $n$ for the 4 SRL configurations: Extension ratio $r_e$, visible extension ratio $r_{e,v}$, and collaboration ratio $r_c$.
    }
    \label{fig:2a:augmentation_quantification}
\end{figure}

\subsection{Reconfiguration Strategy}
% WHY RECONFIGURATION STRATEGY
Adapting SRL configurations to different tasks, as well as to distinct human kinematics, requires a comprehensive strategy. We introduce a strategy that reconfigures the placement $p$ and the number of modules $n$ of the SRL based on the required reachability and task type. The human augmentation analysis provides the basis for this reconfiguration.

To determine the optimal \textbf{placement and configuration} of the Robogami Third Arm for the desired task types, we run the augmentation analysis for our reconfigurable SRL (see `Robogami Third Arm: Design and Capacity' in Methods) when attached to different locations on the body. We select four locations as examples (the chest, back, front of the leg, and side of the leg as seen in Fig. \ref{fig:2a:augmentation_quantification}A) because these attachment scenarios span different combinations of collaborative workspace $C$, visible extended workspace $E_{\text{V}}$, and non-visible extended workspace $E_{\text{NV}}$, thereby capturing a range of augmentation tasks enabled by the SRL.

% Furthermore, these locations can support weight of the arm at the maximum weight (< 1kg). Normally, whole body can be scanned and the optimal locations can be found throught optimization.

We calculate the augmentation ratios (Fig. \ref{fig:2a:augmentation_quantification}B) for these scenarios using the Robogami Third Arm (with module kinematics $M$) as a function of the number of modules $n$. These ratios are computed via a kinematics-based analysis of the human and Robogami Third Arm workspaces, where workspace volumes are estimated using the Robogami Third Arm’s forward kinematics and an average human body model. While we use average anthropometric data for the human model, the same calculation can be adapted to an individual user’s body dimensions. Details on the SRL workspace calculation are presented in the Methods (see `SRL Workspace Calculation'). The augmentation ratios highlight variability and trade-offs in workspace extension and human-SRL collaboration across attachment locations.

\begin{figure}[]
    \centering
    \includegraphics[width=\linewidth]{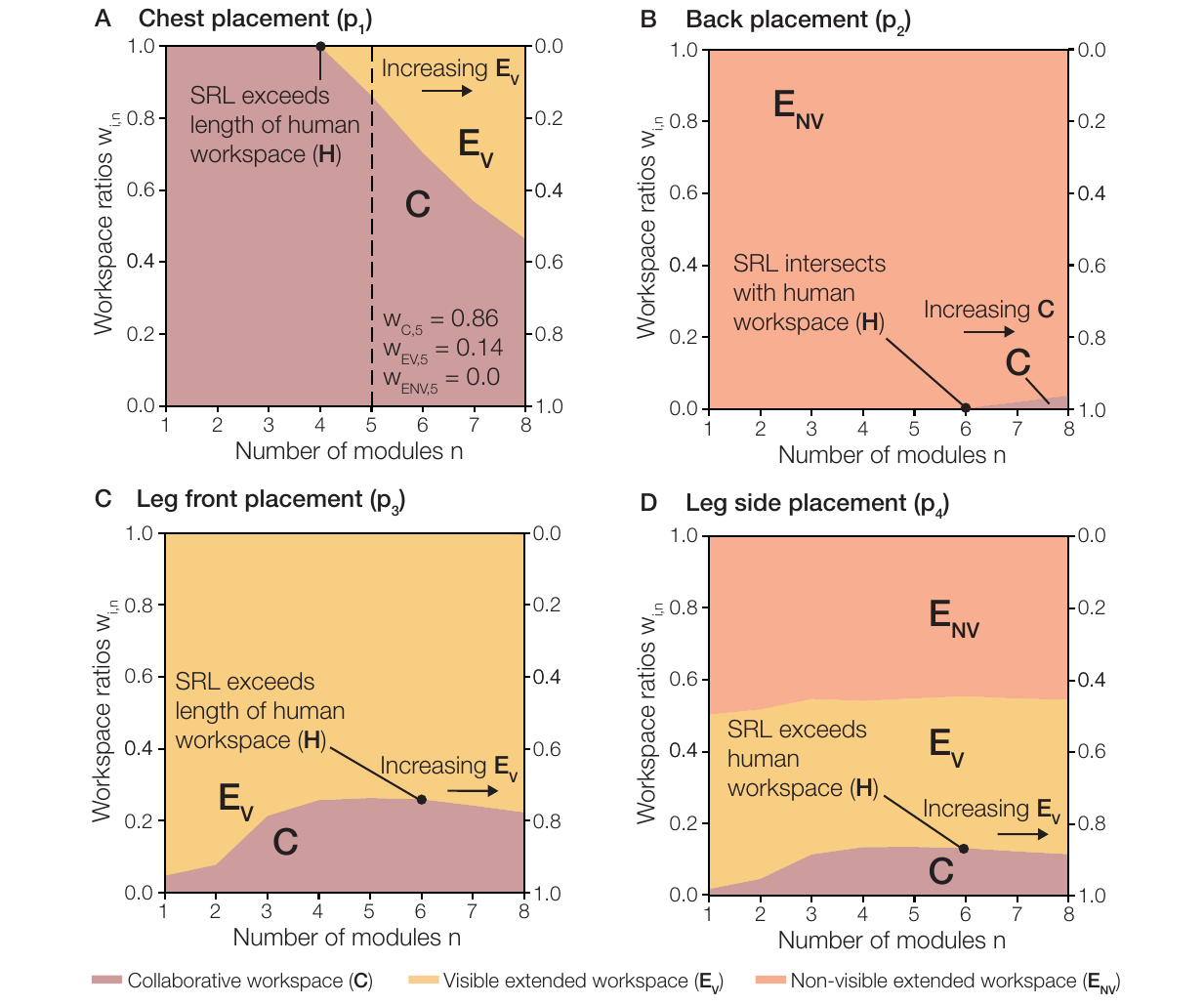}
    \caption{\textbf{Workspace analysis to inform the reconfiguration strategy. } The SRL configuration, defined by the placement $p$ and the number of modules $n$, must be selected according to the scenario and task requirements. Workspace analysis supports this selection by computing workspace ratios $w_{i,n}$ for the collaborative workspace ($C$), the visible extended workspace ($E_{\text{V}}$), and the non-visible extended workspace ($E_{\text{NV}}$). The 4 SRL configurations ($p_1$-$p_4$) on the human body, that capture the range of potential augmentation tasks enabled by the SRL, are analyzed (\textbf{A}-\textbf{D}).}
    \label{fig:2b:workspace_map}
\end{figure}

We compute the workspace ratios, $w_{i,n}$, to create workspace diagrams for the different SRL placements $p_j$ (Fig. \ref{fig:2b:workspace_map}). These diagrams provide the necessary insight to select the minimum number of modules required for the desired task types.

% Chest
For the chest scenario, $p_1$, the workspace ratio for the collaborative workspace $C$ is $w_\text{C}=1.0$ for $n \leq 4$ modules (Fig. \ref{fig:2b:workspace_map}A). For $n > 4$, the Robogami Third Arm length exceeds the human arm length, and the robot workspace $R$ consists of a combination of collaborative workspace and visible extended workspace, $R = C \cup E_{\text{V}}$. As $n$ increases, the workspace ratio for the collaborative workspace, $w_\text{C}$, decreases, while the workspace ratio for the visible extended workspace, $w_{\text{EV}}$, increases. Thus, tasks that require extension of the human workspace require $n > 4$ modules.

% back
For the back scenario, $p_2$, the workspace ratio for the non-visible extended workspace $E_{\text{NV}}$ is $w_{\text{ENV}} = 1.0$ for $n \leq 6$ (Fig. \ref{fig:2b:workspace_map}B). For $n > 6$ modules, the Robogami Third Arm reaches the human arms workspace $H$ and begins to include a collaborative workspace. Thus, tasks that require both workspace extension and collaboration with the human arms require $n > 6$ modules.

% leg front % instead of leg
For placement on the front of the leg, $p_3$, the robot workspace $R$ consists of a combination of collaborative workspace and visible extended workspace, $R = C \cup E_{\text{V}}$. In contrast, for placement on the side of the leg, $p_4$, the robot workspace $R$ includes all three workspace types, $R = C \cup E_{\text{V}} \cup E_{\text{NV}}$. While visual feedback is available across the SRL workspace when placed at $p_3$, it is not available for the $E_{\text{NV}}$ portion when placed at $p_4$.

% SUMMARY (AND TRANSITION TO CONTROL)
In summary, reconfiguration based on the workspace ratios $w_{i,n}$ dictates SRL morphology and placement depending on the desired task and scenario. The human augmentation analysis ensures that reconfiguration decisions are grounded in a quantitative evaluation.

\subsection{Reconfigurable Supernumerary Robotic Limb}
\label{sec:methods:reconfigurable_SRL}
% INTRO WHY
SRL efficacy in dynamic and unstructured environments requires appropriate placement and storage across diverse locations on the human body without hindering the wearer’s movement. Therefore, SRLs should be lightweight, compact, and adaptable in placement and morphology. To address these challenges, we introduce a reconfigurable SRL, referred to as the \textbf{Robogami Third Arm}, which leverages structural and spatial reconfiguration to enhance SRL adaptability (Movie S2).

% WHAT MODULES (SINGLE MODULE)
The Robogami Third Arm consists of origami-inspired robotic modules with 3 DoF \cite{mintchev2019,Robertson2021,mete2021,wang2025}. Each module’s parallel kinematic structure combines the Canfield joint principle with three origami waterbomb legs (Fig. S1). This origami-inspired design offers advantages in lightweight construction and reconfigurability compared with rigid SRLs. With an extension ratio of 3.5 between minimum and maximum SRL length, and a load-to-weight ratio of 2 using two modules, the Robogami Third Arm outperforms rigid SRLs and is comparable to current soft SRLs (Tab. S1). Its weight of \SI{150}{}-\SI{250}{\gram} per module improves wearer comfort and enables placement on various parts of the body, such as the chest, back, or leg. Structural reconfiguration from a flat state to an extended state improves wearability and usability in everyday scenarios. This reconfiguration also enables compact storage of the Robogami Third Arm on the body (Fig. S2).

% \ref{fig:supp:robogami_modules} \ref{tab:supp:related_work_comparison}
%\ref{fig:supp:reconfiguration}

% WHAT MULTIPLE MODULES   
To spatially reconfigure the Robogami Third Arm for adaptation to different scenarios and tasks, we leverage its modularity. Prior research has explored parallel distributions of Robogami modules \cite{salerno2020,jiang2025}; in contrast, this work uses a serial distribution to construct the Robogami Third Arm. By stacking modules in series, we create Robogami arms of variable length (Fig. S3). Overall, the Robogami Third Arm combines structural and spatial reconfiguration through origami-inspired design and modularity to adapt to dynamic and unstructured environments. Further details on the Robogami Third Arm design and capacity are provided in `Robogami Third Arm: Design and Capacity' in Methods.
% \ref{fig:supp:modularity}

\subsection{Adaptable Control Strategy}
% WHY IS ADAPTABLE CONFIGURATION IMPORTANT
Effective task execution, either independently or in coordination with the human’s natural limbs, requires a control strategy that adapts to varying SRL configurations, scenarios, and tasks. Therefore, we introduce an adaptable control strategy that accounts for the SRL’s number of modules $n$, attachment location $p$, and kinematic constraints.

\begin{figure}[h!]
    \centering
    \includegraphics[width=\linewidth]{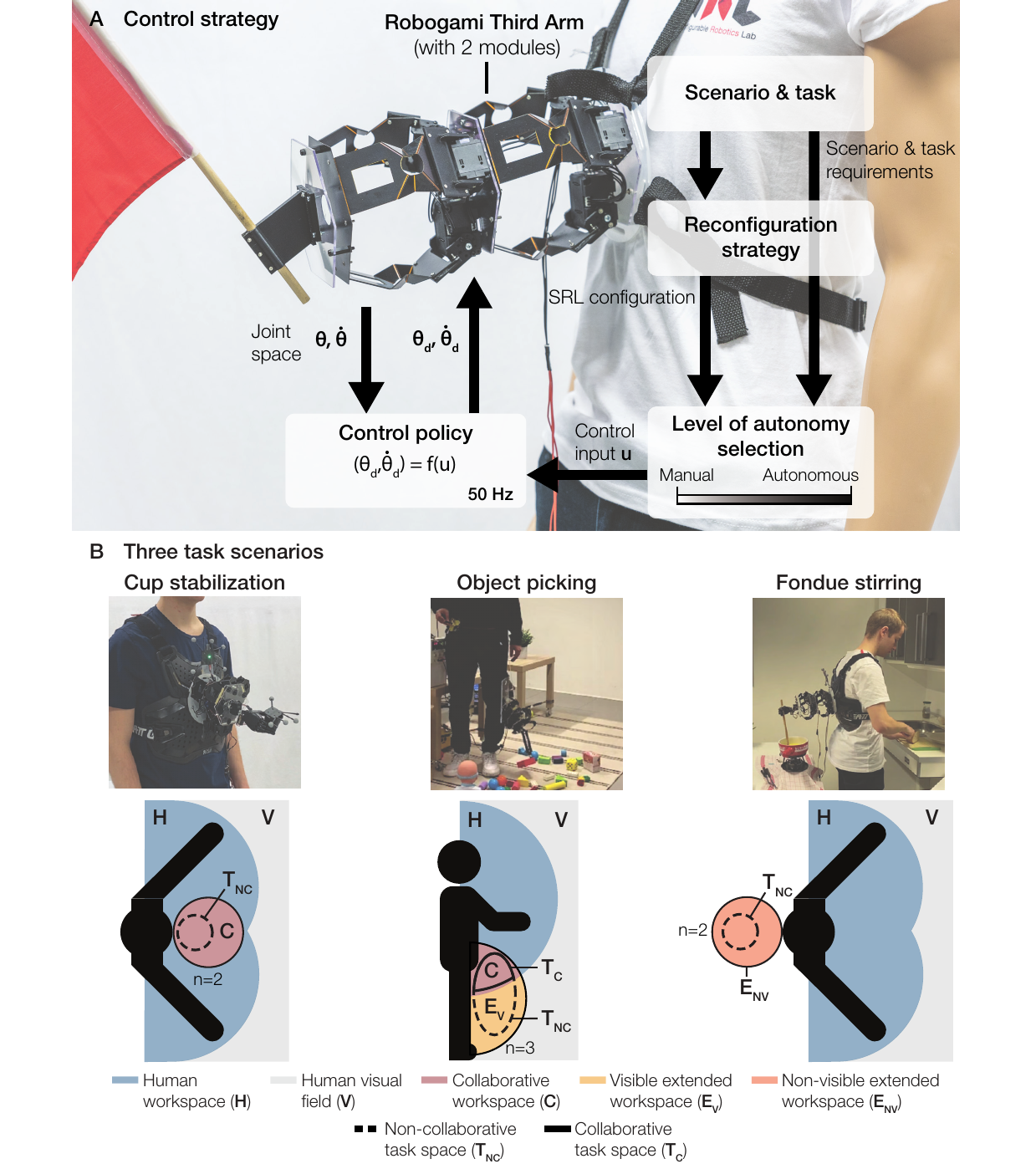}
    \caption{\textbf{Adaptable control strategy. }(\textbf{A}) Overview of the adaptable control strategy and its main components: the reconfiguration strategy to determine the SRL configuration based on scenario and task requirements; the selection of the appropriate autonomy level; and the control policy to control the Robogami Third Arm. (\textbf{B}) Three exemplary task scenarios demonstrating the reconfiguration and adaptable control strategy for selecting the SRL configuration and autonomy level.}
    \label{fig:3:adaptable_control}
\end{figure}

% WHAT SCALABLE CONTROL STRATEGY ARE WE INTRODUCING FOR RECONFIGURABLE SRL
This strategy consists of three components (Fig. \ref{fig:3:adaptable_control}A): (i) the reconfiguration strategy, (ii) level-of-autonomy selection, and (iii) the control policy. The reconfiguration strategy and the level-of-autonomy selection use the quantified human augmentation analysis together with task requirements to determine the SRL configuration and an appropriate autonomy level. Finally, the control policy uses the control-interface input $u$ to generate and execute joint trajectories $(\theta_d,\dot{\theta}_d)$ for the SRL.

% WHAT LEVEL OF AUTONOMY TO SELECT AND HOW
To select the \textbf{level of autonomy}, we leverage the human augmentation analysis together with task requirements. We divide tasks into collaborative and non-collaborative categories with task spaces $T_C$ and $T_{NC}$, respectively, whose union defines the full task space $T = T_C \cup T_{NC}$. For collaborative tasks, the SRL must coordinate closely with the natural limbs, requiring high-fidelity intention detection; thus, manual control is preferred. Non-collaborative tasks can be executed primarily by the SRL, enabling multitasking. For these tasks, either manual or autonomous control can be used depending on the availability of sensory feedback and the user’s preference.

Based on the augmentation analysis, the user selects a suitable body position and reconfigures the SRL by adjusting its attachment location $p$ and number of modules $n$ to satisfy two conditions: (i) the collaborative task space must lie within the collaborative workspace, and (ii) the overall task space must lie within the augmented workspace. If no SRL configuration satisfies these conditions, the user must change position relative to the task and repeat the search for a feasible configuration:
\begin{equation}
T_C \subset C \quad \mathrm{and} \quad T \subset A
\end{equation}

Given a feasible configuration, we select the autonomy level based on task type and sensory feedback. If the task is collaborative and $T_C \subset C$, we select manual control because high intention detection is required. If the task is non-collaborative and lies in the non-visible extended workspace, $T_{NC} \subset E_{\text{NV}}$, we select autonomous control because sensory feedback is unavailable and explicit intention detection is not required. In the remaining cases, both manual and autonomous control are feasible; the user can bias toward autonomous control to reduce cognitive load or toward manual control to increase intentionality and the sense of embodiment.

To demonstrate the proposed reconfiguration and adaptable control strategies, we consider three task scenarios (Fig. \ref{fig:3:adaptable_control}B): (i) holding and stabilizing a coffee cup while manipulating other objects with the natural hands, (ii) picking up objects from the floor and transferring them to the natural hands, and (iii) stirring a Swiss fondue while chopping other ingredients with the natural hands (Movie S3). While SRLs can support many tasks, these scenarios span different task types and task-space requirements, enabling validation of the proposed strategy.

Stabilizing a cup is a non-collaborative task and can be performed without interfering with natural-arm actions. We attach the Robogami Third Arm with $n=2$ modules to the chest. This configuration allows the SRL to hold the cup steady (e.g., to prevent spilling) and to position it within reach of the user’s mouth if desired. For this placement, the robot workspace is dominated by the collaborative workspace, with $w_{\text{max}}=w_\text{C}=1.0$. Because the configuration lies within the collaborative workspace, it also permits transferring the cup to a natural hand. Although both manual and autonomous control are feasible, we use autonomous control in this scenario to reduce cognitive load while enabling multitasking.

Picking objects from the floor and transferring them to the natural hands is a collaborative task that requires coordination with the user’s arms. Because the task space lies below the body, we place the SRL on the front of the leg, which provides favorable coverage of the lower workspace with a small number of modules, thereby maintaining a higher payload capability. We attach the Robogami Third Arm with $n=3$ modules to the leg, resulting in $w_\text{C}=0.21$ and $w_{\text{max}}=w_{\text{EV}}=0.79$. Since the task requires collaboration, we use manual control through a custom joystick interface, implemented as a miniaturized and sensorized version of the Robogami modules (Fig. S4, 'Robogami Third Arm: Design and Capacity' in Methods).
% \ref{fig:supp:user_interfaces}

Stirring while simultaneously preparing ingredients is non-collaborative and is primarily intended to enable multitasking. We mount the SRL on the back to spatially separate the SRL task space from the natural-hand task space, reducing the risk of interference. Using $n=2$ modules on the back yields a robot workspace dominated by the non-visible extended workspace, with $w_{\text{max}}=w_{\text{ENV}}=1.0$. We therefore use autonomous control for this scenario due to the limited visual feedback.

In summary, the adaptable control strategy couples SRL reconfiguration with autonomy selection by mapping task requirements to the augmented workspace classification. By ensuring task feasibility through $T_C \subset C$ and $T \subset A$, and selecting manual or autonomous control based on collaboration needs and sensory feedback availability, the strategy enables task-dependent SRL operation across placements and configurations.

\subsection{Control Policy Evaluation}

% WHAT CONTROL POLICY FOR ANY NUMBER OF MODULES AND HOW (QP)
Kinematic modeling of the Robogami Third Arm becomes significantly more complex as the number of modules, $n$, increase, making it challenging to derive a closed-form inverse-kinematics solution required to generate robot motions. Additionally, due to kinematic redundancy, the system often presents infinitely many joint configurations that achieve a desired end-effector pose. Consequently, purely model-based inverse-kinematics approaches are not scalable with increasing $n$ and end-effector DoF requirements. To address these scalability issues while accounting for internal and external constraints, we employ a \textbf{quadratic-programming (QP)-based control policy} that explicitly incorporates the number of modules $n$ and the kinematic constraints of reconfigurable SRLs. QP-based control has proven effective for real-time control of multi-robot systems and reconfigurable robots with changing morphology \cite{bouyarmane2018quadratic, bolotnikova2025}. The proposed formulation jointly handles internal constraints, such as closed-chain kinematics and folding hinge limits, as well as external constraints, such as self-collision avoidance, while producing joint-level position and velocity commands. The QP-based controller accepts commands from either autonomous motion planners or manual control interfaces and computes the required joint motion, which is then tracked by a PID-based joint position and velocity controller (Fig. S5). Further details on the QP-based formulation are provided in `Quadratic Programming-based Control Policy: Formulation' in Methods.
% \ref{fig:supp:low_level_joint_control}

\begin{figure}[t]
    \centering
    \includegraphics[width=\linewidth]{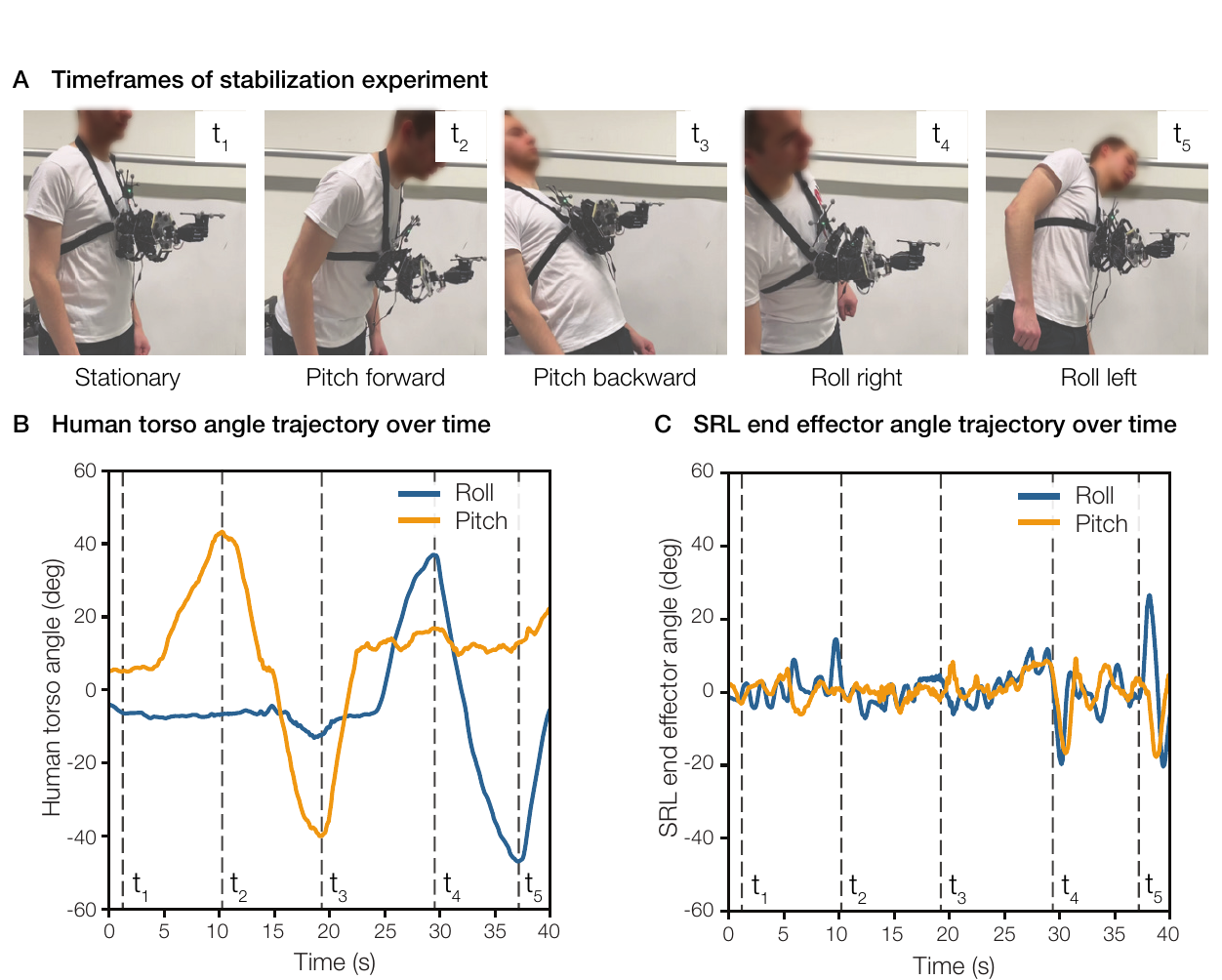}
    \caption{\textbf{Control policy evaluation through object stabilization. }The closed-loop performance of the control policy is evaluated using an object stabilization experiment (cup holding). (\textbf{A}) Representative time steps from the human-in-the-loop stabilization experiment. The participant applies disturbances to the end effector in pitch (t2 and t3) and roll (t4 and t5). (\textbf{B}) Torso angle of the participant over time, indicating the applied disturbances during the experiment. (\textbf{C}) End-effector angle of the SRL, the Robogami Third Arm, over time in response to the human-applied disturbances.}
    \label{fig:5:cup_stabilization_results}
\end{figure}

% WHY THIS EXPERIMENT
We validate the closed-loop control performance of the proposed policy through an object stabilization experiment (Fig. \ref{fig:5:cup_stabilization_results}, Movie S4). The goal is to maintain the end-effector orientation parallel to the ground under external disturbances. This experiment resembles scenarios in which an SRL holds a cup containing liquid and must prevent spilling while the human wearer moves.

% WHAT EXPERIMENTAL SETUP
For this experiment, we formulate the QP-based control policy with a multi-objective function that aims to minimize the error between the end effector orientation and a world-frame orientation set on the ground. We utilize a motion capture system (Vicon Vero, USA) for tracking the human torso orientation, the Robogami Third Arm's end effector orientation, and the world frame orientation set on the ground. The Robogami Third Arm is worn by a human subject to create external disturbances through rotations around the human torso (Fig. \ref{fig:5:cup_stabilization_results}A).

% WHAT RESULTS 
The human subject rotated their torso with a pitch angle of approximate \SI{40}{\degree} in positive (t2) and negative rotation direction (t3), as well as a roll angle of approximate \SI{40}{\degree} in positive (t4) and negative rotation direction (t5) (Fig. \ref{fig:5:cup_stabilization_results}B). During the human disturbances, the absolute error of the end effector roll angle is \SI{4.2 \pm 4.5}{\degree} and reaches a maximum absolute angle of \SI{27}{\degree} (Fig. \ref{fig:5:cup_stabilization_results}C). The absolute error of the pitch angle is \SI{3.0 \pm 3.4}{\degree} and reaches a maximum absolute angle of \SI{18}{\degree}. 

% SUMMARY
In summary, the end effector stabilization results show that the QP-based control policy is able to effectively control the Robogami Third Arm with external disturbances, such as movements of the human wearer, while taking into account the number of modules $n$ and the kinematic constraints of the Robogami Third Arm. The experiments highlight the potential of QP-based control for achieving safe and complex task execution with reconfigurable SRLs.

\section{Discussion}
This study advances SRLs toward operation in dynamic, unstructured settings by coupling a quantitative augmentation model with reconfiguration and adaptable control. We introduce a human augmentation analysis that quantifies how SRL morphology, placement, and the wearer's available sensory feedback shape the augmented workspace, with vision as a primary channel. By classifying the SRL workspace into collaborative, visible extended, and non-visible extended regions, the framework links where the SRL can act to how it should be controlled. This workspace classification enables systematic comparisons across configurations using augmentation and workspace ratios, providing a quantitative basis for selecting SRL placement, length, and autonomy in a task-dependent manner.

To demonstrate our approach, we present a compact, lightweight Robogami Third Arm built from origami-inspired modular units with intrinsic compliance. This design supports comfortable wearability across multiple on-body locations. Structural reconfiguration from a flat to an extended state enables unobtrusive on-body storage and rapid deployment, while serial modular stacking provides variable reach without redesigning the device. For control, we employ a quadratic-programming-based policy that remains scalable as the number of modules increases and that explicitly incorporates internal and external constraints. Using the integrated reconfiguration and autonomy-selection strategy, we demonstrate task-dependent operation across three representative scenarios: coffee cup stabilization with the SRL mounted on the chest, pick-and-transfer of objects from the ground with the SRL mounted on the leg, and Swiss fondue stirring with the SRL mounted on the back.

Several limitations motivate future work. First, our sensory-feedback modeling emphasizes vision; while this captures a dominant feedback pathway, additional modalities such as haptic cues, auditory feedback, or camera-based displays could expand the effective feedback workspace and shift the boundary between regions best suited for manual versus autonomous control. Second, our validation focuses primarily on reachability-driven tasks; the analysis could be extended to incorporate payload limits, contact conditions, and task DoF requirements. Third, our workspace analysis and autonomy-selection rules are evaluated under assumptions about user pose/position and quasi-static postures during planning. Future work will extend the framework to real-time, dynamic scenarios in which both the wearer and the task evolve over time.

Overall, this work provides a quantitative approach to SRLs that can be configured and controlled for unstructured, everyday tasks. By explicitly linking SRL placement and morphology to workspace utility, feedback availability, and autonomy requirements, the proposed framework moves SRLs beyond task-specific deployments toward adaptable augmentation that can be deployed where and when assistance is needed.

\section{Methods}

\subsection*{SRL Workspace Calculation}
\label{sec:methods:workspace_calculation}
% WHY ARE WE DOING THE HUMAN-ROBOT WORKSPACE ANALYSIS AND HOW ARE WE DOING THE WORKSPACE ANALYSIS
We calculate the workspaces for varying SRL configuration to perform the human augmentation analysis by utilizing the kinematic model of a human model and the SRL, as well as the human visual field model (Fig. S6). Based on the models, we compute point clouds for the human and robot workspace ($H$ and $R$) and the point cloud constraints for the human visual field space ($V$). We utilize the boolean equations (Equation \ref{eq:workspace_calculation}) for the collaborative workspace, $C$, the visible extended workspace, $E_{\text{V}}$, and the non-visible extended workspace, $E_{\text{NV}}$ to compute the point cloud for each workspace. The point clouds are then transformed into a convex polytope to calculate the volume of each workspace.
% \ref{fig:supp:adaptable_configuration}

% WHAT DID WE DO IN OUR CASE
For the evaluations, we use a human kinematic model with 4 DoF for each arm. These DoF include shoulder movements with 3 DoF and elbow movements with 1 DoF. Based on the functional range of motion for the elbow \cite{sardelli2011} and the shoulder \cite{namdari2012}, we define the joint limits for the kinematic model:
\begin{equation}
    \begin{split}
    \theta_{\text{elbow}} \in [\SI{0}{\degree}, \SI{130}{\degree}], \\
    \theta_{\text{shoulder, adduction}} \in [\SI{-90}{\degree}, \SI{25}{\degree}], \\
    \theta_{\text{shoulder, abduction}} \in [\SI{0}{\degree}, \SI{130}{\degree}], \\
    \theta_{\text{shoulder, flexion-extension}} \in [\SI{0}{\degree}, \SI{120}{\degree}].
    \end{split}
\end{equation}
We based the bone length of the human kinematic model on average human body data taken from the NASA anthropometric source book \cite{churchill1978}. However, the calculations can be adapted to the user's specific body size. The human visual field space, V, is defined as the space in front of the human. 
We utilize the analytical model presented in prior work \cite{mete2021} to obtain the forward kinematic model $f_{i}(\boldsymbol{\theta})$ of the Robogami module $i$ as a transformation matrix $T_{i}(\boldsymbol{\theta})$. Based on matrix multiplication, we obtain the forward kinematic model of a complete Robogami Third Arm $T_{\text{arm}}$ with $n$ modules and the motor joint limits:
\begin{equation}
    T_{\text{arm}} = \prod_{i=1}^{n}  T_{i}(\boldsymbol{\theta}) : \boldsymbol{\theta} \in [\SI{5}{\degree}, \SI{85}{\degree}].
\end{equation}
Based on the Robogami Third Arm kinematic model, the Robogami Third Arm workspace can be calculated as a function of the number of modules $n$ used. We uniformly sample the joint space with $10000$ samples for the human and $10000$ samples for the Robogami Third Arm. For the human visual field space, we use the full space in front of the human, which are realized as positional constraints for the point clouds. We use k-dimensional trees to compute the intersections and differences between the point clouds. Subsequently, we convert the computed point clouds into convex polytope meshes to compute the volume of the different workspaces. The thresholds for the k-dimensional trees are selected, so that the reconstructed robot volume has a maximum difference of $10$ $\%$ to the actual robot volume. 
%We use the Open3D library to compute the point clouds and convex polytope meshes \cite{Zhou2018}.
%Using this method, we are able to calculate the workspaces for any number of modules $n$ and any placement on the human body for adapting the Robogami Third Arm to various tasks and scenarios.
% Using the human and robot kinematic model, we compute the collaborative workspace, the extended workspace inside the human's vision, and the extended workspace outside of the human's vision as a function of the Robogami Third Arm's number of modules $n$ (Fig. \ref{fig:supp:adaptable_configuration}). 

\subsection*{Robogami Third Arm: Design and Capacity}
\label{sec:methods:robogami}
% WHAT DESIGN ARE WE USING FOR THE Robogami Third Arm AND HOW ARE WE MANUFACTURING THE Robogami Third Arm
The Robogami Third Arm, a reconfigurable SRL, comprises multiple modules with identical kinematics (Fig. S1). Each module utilizes 3 origami waterbomb legs that are connected together to build the parallel kinematic structure, similar to the structures used in prior work \cite{mintchev2019, Robertson2021, mete2021, wang2025}. The origami legs are a sandwich structure based on two rigid FR4 layers and a flexible polyimide layer in between, which are connected to each other using two adhesive layers activated through heat pressing. The legs enable the reconfiguration of the structure through actuation of motors (Dynamixel, Robotis) connected to each leg. We utilize motors with a stall torque of \SI{1.4}{\N\m} for modules with higher payload and motors with a stall torque of \SI{0.52}{\N\m} for lightweight modules with lower payload. 
% \ref{fig:supp:robogami_modules}

% WHY PAYLOAD CAPACITY CHARACTERIZATION
%The payload capacity of reconfigurable SRLs, consisting of multiple modules, is dependent on the number of modules used.

% WHAT EXPERIMENT TO COMPARE THE MODEL
To quantify the payload capacity of the Robogami Third Arm, we conduct payload experiments using the arm (with 1 to 3 modules) in a static, fully extended, straight-beam configuration (Fig. S7). This configuration represents the most challenging orientation for the base motors, as it corresponds to the arm’s maximum extension and therefore generates the highest torque from the combined weight of the arm and the payload.
%which is the most challenging orientation for the motors at the base to carry the torque generated by the arm's weight and payload. 
We record the maximum payload of the Robogami Third Arm by increasing the attached weights until the maximum torque is reached. 
The end condition for the tests are one of the motors reach to maximum torque output cannot keep the arm straight or the arm mechanically fails. The experiments with straight-beam load case show a payload capacity of \SI{3000}{\gram} for 1 module, \SI{1000}{\gram} for 2 modules, and \SI{300}{\gram} for 3 modules, while each module weighs \SI{150}{}-\SI{250}{\gram}.

% \ref{fig:supp:payload_capacity}
%After reaching the maximum torque, large deflection of the Robogami Third Arm structure are damaging the flexible folding-based joints of the origami legs.

% HOW ARE WE CONTROLLING THE ARM
The Robogami Third Arm's central controller unit (Arduino Due) performs the low-level joint control of the motors, coordinates the communication from and to the high-level control, and manages the current flow from the power supply (Fig. S8). The high-level QP-based controller runs on an external computer and sends the desired values to the central controller, where the low-level joint control moves the motors.
Wireless communication with peripheral sensors, such as manual control interfaces, is achieved via Bluetooth Low Energy (BLE) using a wireless ESP32 module connected to the central controller unit. For manual control of the Robogami Third Arm, we design a 3-DoF joystick based on a sensorized miniaturized version of the arm (Fig. S4). The joystick uses potentiometers to measure joint angles, enabling intuitive manual operation.
% \ref{fig:supp:communication}
% \ref{fig:supp:user_interfaces}

% HOW DO WE RUN THE Robogami Third Arm UNTETHERED
Finally, we designed a compact box for housing the power (6S lithium-polymer battery), control (Arduino Due), and communication peripherals (ESP32) to operate the Robogami Third Arm untethered on the human body (Fig. S9).
% \ref{fig:supp:untethered_box}
%A  provides the required power for operating the robot arm. 
%The motors are connected with each other and the central controller unit (Arduino Due with Dynamixel Shield) using TTL communication. 
%The  
%Using a DC/DC converter, we provide the necessary voltage for the motors (\SI{12}{\volt}) and the microcontrollers (\SI{3.3}{\volt}).
%The Dynamixel motors are controlled with an Arduino Due and a matching Arduino shield for Dynamixel motors. 
%In our case, we utilize Bluetooth Low Energy (BLE) for the wireless communication to the sensors.
%For enabling communication with the QP-based high level controller, we utilize Serial communication between the Arduino Due and an external computer.

\subsection*{Quadratic Programming-based Control Policy: Formulation} 
% HOW DOES THE QP BASED CONTROL ACTUALLY WORK
To generate the motion for reconfigurable SRL with varying configurations, the QP-based control policy for autonomous control comprises three main components: SRL module description, the control objective function, and control constraints.

% MODULE DESCRIPTION
The SRL module description describes the SRL module's structure and consists of a kinematic tree that details the dimensions of all module links and their interconnections via joints. For the Robogami Third Arm, the origami-inspired waterbomb joints are modeled as three revolute joints with the same origin. Based on 
%Based on five main parameters (leg length, leg thickness, base side length, base thickness, and waterbomb joint angle) the kinematic description of the Robogami module is created.
%These parameters are entered into a script that generates a Universal Robot Description File (URDF) containing the complete kinematic description of the module structure. 
%If necessary, this URDF file can be manually post-processed to add auxiliary virtual frame descriptions (e.g., for the custom end-effector position) to facilitate the definition of control objectives.

% CONTROL OBJECTIVE FUNCTION
The control objective function describes the control task goal and consists of several weighted objectives or tasks. The primary task, always present in the objective function, is called the posture task, which specifies the preferred position for each joint of the module and ensures that the optimization problem is well-posed. The posture task typically has a low-weight in the objective function compared to other tasks. To represent the folding joints of the Robogami Third Arm, different weights are assigned to passive and active joints in the posture task. The weights assigned to passive joints are significantly lower than those assigned to the three main active joints of the Robogami module, ensuring that the QP solution is not significantly influenced by the posture task values of the passive joints.
The second task is the end effector position and orientation task associated with an end effector point of the outermost module. Additional tasks may be included depending on the task and scenario, such as position or image-based visual servoing tasks for vision-based closed-loop end-effector control or trajectory tracking tasks.

% CONSTRAINTS
The control constraints describe the constraints to ensure safe control of the reconfigurable SRL.
% Module constraints
For the Robogami Third Arm, four constraints for each module ensure the proper representation of the module and its kinematic capabilities in the control problem. 
First, two internal closed-loop fixed contact constraints to represent the closed-chain kinematics of the Robogami module \cite{mintchev2019, Robertson2021, mete2021, wang2025}, as well as one virtual joint or fixed contact constraint with one free axis to allow the motion of a passive joint around the axes. 
The second constraint considers the active joint limits in terms of position, velocity, and acceleration for the three main active (controlled) joints of a module. 
The third constraint considers the motion limits of the origami-inspired folding hinges to prevent overstretching and braking of the hinges. 
The fourth constraint considers self-collision avoidance to prevent the Robogami module from colliding with itself. 
% complete SRL
In addition to the constraints for a single SRL module, fixed contact constraints between multiple SRL modules are defined. These constraints can be added or removed from the control problem formulation during runtime in case of reconfiguration of the Robogami Third Arm.

\backmatter

\bmhead{Supplementary information}
The supplementary information is available in a separate file.

\bmhead{Acknowledgements}
We thank Kevin Holdcroft for his contributions to the design and fabrication of the wearable power supply unit for untethered operation of the Robogami Third Arm. We also thank Prof. Jingjing Li (University of Tsukuba) for her illustration work on Fig. \ref{fig:1:vision}. The image in Fig. \ref{fig:3:adaptable_control} is courtesy of Jamani Caillet (EPFL Mediacom). This work was supported by the Swiss National Science Foundation under the grant “Approaching Tangible Reality” (grant no. 204819). The study was approved by the Human Research Ethics Committee at EPFL (No. 096-2023).

\bmhead{Author contributions}
Conceptualization M.M., A.B., A.S., and J.P.; Funding acquisition J.P.; Investigation M.M., A.B., and A.S.; Methodology M.M., A.B., and A.S.; Software M.M., A.B. and A.S.; Supervision J.P.; Visualization M.M., A.B. and A.S.; Writing - original draft M.M., A.B. and A.S.; Writing - review \& editing M.M., A.B., A.S., and J.P.

\bmhead{Competing interests}
The authors declare no competing interests.

\bmhead{Correspondence} Correspondence and requests for materials should be addressed to Jamie Paik.

\bibliography{sn-bibliography}% common bib file

@article{Parietti2014_1,
   author = {Federico Parietti and H Harry Asada},
   journal = {IEEE International Conference on Robotics and Automation (ICRA)},
   title = {Supernumerary Robotic Limbs for Aircraft Fuselage Assembly: Body Stabilization and Guidance by Bracing},
   year = {2014},
}

@article{Parietti2014_2,
   author = {Federico Parietti and Kameron Chan and H Harry Asada},
   journal = {IEEE International Conference on Robotics and Automation (ICRA)},
   title = {Bracing the Human Body with Supernumerary Robotic Limbs for Physical Assistance and Load Reduction},
   year = {2014},
}

@article{Ballesteros2023,
   author = {Erik Ballesteros and Brandon Man and H Harry Asada},
   journal = {IEEE International Conference on Robotics and Automation (ICRA 2023)},
   title = {Supernumerary Robotic Limbs for Next Generation Space Suit Technology},
   year = {2023},
}

@article{veronneau2020,
  title={Multifunctional remotely actuated 3-DOF supernumerary robotic arm based on magnetorheological clutches and hydrostatic transmission lines},
  author={V{\'e}ronneau, Catherine and Denis, Jeff and Lebel, Louis-Philippe and Denninger, Marc and Blanchard, Vincent and Girard, Alexandre and Plante, Jean-S{\'e}bastien},
  journal={IEEE Robotics and Automation Letters},
  volume={5},
  number={2},
  pages={2546--2553},
  year={2020},
  publisher={IEEE}
}

@article{Zhang2022,
   author = {Qinghua Zhang and Changle Li and Hongwei Jing and Hongwu Li and Xianglong Li and Haotian Ju and Yuan Tang and Jie Zhao and Yanhe Zhu},
   issue = {3},
   journal = {IEEE Robotics AND Automation Letters},
   title = {RTSRAs: A Series-Parallel-Reconfigurable Tendon-Driven Supernumerary Robotic Arms},
   volume = {7},
   year = {2022},
}

@article{Zhang2023,
   author = {Chao Zhang and Zhuang Zhang and Yun Peng and Yanlin Zhang and Siqi An and Yunjie Wang and Zirui Zhai and Yan Xu and Hanqing Jiang},
   issue = {1},
   journal = {Nature Communications},
   month = {12},
   title = {Plug \& play origami modules with all-purpose deformation modes},
   volume = {14},
   year = {2023},
}

@article{Robertson2021,
   author = {Matthew A. Robertson and Özdemir Can Kara and Jamie Paik},
   issue = {1},
   journal = {International Journal of Robotics Research},
   month = {1},
   pages = {72-85},
   title = {Soft pneumatic actuator-driven origami-inspired modular robotic “pneumagami”},
   volume = {40},
   year = {2021},
}

@article{Nguyen2019_1,
   author = {Pham Huy Nguyen and Curtis Sparks and Sai G. Nuthi and Nicholas M. Vale and Panagiotis Polygerinos},
   issue = {1},
   journal = {Soft Robotics},
   month = {2},
   pages = {38-53},
   title = {Soft Poly-Limbs: Toward a New Paradigm of Mobile Manipulation for Daily Living Tasks},
   volume = {6},
   year = {2019},
}

@article{Kusunoki2023,
   author = {Mikiya Kusunoki and Linh Viet Nguyen and Hsin-Ruey Tsai and Van Anh Ho and Haoran Xie},
   journal = {IEEE Access},
   title = {Scalable and Foldable Origami-Inspired Supernumerary Robotic Limbs for Daily Tasks},
   year = {2023},
}

@article{Nguyen2019_2,
   author = {Pham H. Nguyen and Imran I. B. Mohd and Curtis Sparks and Fracisco L. Arellano and Wenlong Zhang and Panagiotis Polygerinos},
   journal = {International Conference on Robotics and Automation (ICRA)},
   title = {Fabric Soft Poly-Limbs for Physical Assistance of Daily Living Tasks},
   year = {2019},
}

@article{Liu2021,
   author = {Sicong Liu and Yuming Zhu and Zicong Zhang and Zhonggui Fang and Jiyong Tan and Jing Peng and Chaoyang Song and H. Harry Asada and Zheng Wang},
   issue = {5},
   journal = {IEEE/ASME Transactions on Mechatronics},
   month = {10},
   pages = {2747-2757},
   title = {Otariidae-Inspired Soft-Robotic Supernumerary Flippers by Fabric Kirigami and Origami},
   volume = {26},
   year = {2021},
}

@article{dominijanni2023human,
author = {Giulia Dominijanni  and Daniel Leal Pinheiro  and Leonardo Pollina  and Bastien Orset  and Martina Gini  and Eugenio Anselmino  and Camilla Pierella  and Jérémy Olivier  and Solaiman Shokur  and Silvestro Micera },
title = {Human motor augmentation with an extra robotic arm without functional interference},
journal = {Science Robotics},
volume = {8},
number = {85},
pages = {eadh1438},
year = {2023}
}

@article{Kieliba2021,
author = {Paulina Kieliba  and Danielle Clode  and Roni O. Maimon-Mor  and Tamar R. Makin },
title = {Robotic hand augmentation drives changes in neural body representation},
journal = {Science Robotics},
volume = {6},
number = {54},
pages = {eabd7935},
year = {2021}}

@article{bouyarmane2018quadratic,
  title={Quadratic programming for multirobot and task-space force control},
  author={Bouyarmane, Karim and Chappellet, Kevin and Vaillant, Joris and Kheddar, Abderrahmane},
  journal={IEEE Transactions on Robotics},
  volume={35},
  number={1},
  pages={64--77},
  year={2018}
}

@inproceedings{ballesteros2024,
  title = {Supernumerary {{Robotic Limbs}} to {{Support Post-Fall Recoveries}} for {{Astronauts}}},
  booktitle = {{{IEEE International Conference}} on {{Robotics}} and {{Automation}} ({{ICRA}})},
  author = {Ballesteros, Erik and Lee, Sang-Yoep and Carpenter, Kalind C. and Harry Asada, H.},
  year = {2024},
  pages = {2324--2331}
}

@article{clode2024,
  title = {Evaluating Initial Usability of a Hand Augmentation Device across a Large and Diverse Sample},
  author = {Clode, Dani and Dowdall, Lucy and {da Silva}, Edmund and Sel{\'e}n, Klara and Cowie, Dorothy and Dominijanni, Giulia and Makin, Tamar R.},
  year = {2024},
  journal = {Science Robotics},
  volume = {9},
  number = {90},
  pages = {eadk5183}
}

@article{eden2022,
  title = {Principles of Human Movement Augmentation and the Challenges in Making It a Reality},
  author = {Eden, Jonathan and Br{\"a}cklein, Mario and Ib{\'a}{\~n}ez, Jaime and Barsakcioglu, Deren Yusuf and Di Pino, Giovanni and Farina, Dario and Burdet, Etienne and Mehring, Carsten},
  year = {2022},
  month = mar,
  journal = {Nature Communications},
  volume = {13},
  number = {1},
  pages = {1345},
}

@article{goldberg2024,
  title = {Augmented Dexterity: {{How}} Robots Can Enhance Human Surgical Skills},
  shorttitle = {Augmented Dexterity},
  author = {Goldberg, Ken and Guthart, Gary},
  year = {2024},
  month = oct,
  journal = {Science Robotics},
  volume = {9},
  number = {95},
  pages = {eadr5247},
}

@article{lisiniBaldi2025,
  title = {Exploiting Body Redundancy to Control Supernumerary Robotic Limbs in Human Augmentation},
  author = {Lisini Baldi, Tommaso and D'Aurizio, Nicole and Gaudeni, Chiara and Gurgone, Sergio and Borzelli, Daniele and {d'Avella}, Andrea and Prattichizzo, Domenico},
  year = {2025},
  month = feb,
  journal = {The International Journal of Robotics Research},
  volume = {44},
  number = {2},
  pages = {291--316},
}

@article{parietti2016,
  title = {Supernumerary {{Robotic Limbs}} for {{Human Body Support}}},
  author = {Parietti, Federico and Asada, Harry},
  year = {2016},
  journal = {IEEE Transactions on Robotics},
  volume = {32},
  number = {2},
  pages = {301--311},
}

@article{prattichizzo2021,
  title = {Human Augmentation by Wearable Supernumerary Robotic Limbs: Review and Perspectives},
  shorttitle = {Human Augmentation by Wearable Supernumerary Robotic Limbs},
  author = {Prattichizzo, Domenico and Pozzi, Maria and Lisini Baldi, Tommaso and Malvezzi, Monica and Hussain, Irfan and Rossi, Simone and Salvietti, Gionata},
  year = {2021},
  month = sep,
  journal = {Progress in Biomedical Engineering},
  volume = {3},
  number = {4},
  pages = {042005},
}

@article{yang2021,
  title = {Supernumerary {{Robotic Limbs}}: {{A Review}} and {{Future Outlook}}},
  shorttitle = {Supernumerary {{Robotic Limbs}}},
  author = {Yang, Bo and Huang, Jian and Chen, Xinxing and Xiong, Caihua and Hasegawa, Yasuhisa},
  year = {2021},
  journal = {IEEE Transactions on Medical Robotics and Bionics},
  volume = {3},
  number = {3},
  pages = {623--639}
}

@article{bergamasco2016,
  title={Human--robot augmentation},
  author={Bergamasco, Massimo and Herr, Hugh},
  journal={Springer handbook of robotics},
  pages={1875--1906},
  year={2016},
  publisher={Springer}
}

@inproceedings{franco2021,
  title={A manually actuated robotic supernumerary finger to recover grasping capabilities},
  author={Franco, Leonardo and Prattichizzo, Domenico and Salvietti, Gionata},
  booktitle={IEEE International Humanitarian Technology Conference (IHTC)},
  pages={1--4},
  year={2021},
  organization={IEEE}
}

@inproceedings{treers2017,
  title = {Design and Control of Lightweight Supernumerary Robotic Limbs for Sitting/Standing Assistance},
  booktitle = {International Symposium on Experimental Robotics},
  author = {Treers, Laura and Lo, Roger and Cheung, Michael and Guy, Aymeric and Guggenheim, Jacob and Parietti, Federico and Asada, Harry},
  year = {2017},
  pages = {299--308},
  isbn = {978-3-319-50115-4}
}

@article{haoSupernumeraryRoboticLimbs2020,
  title = {Supernumerary {{Robotic Limbs}} to {{Assist Human Walking With Load Carriage}}},
  author = {Hao, Ming and Zhang, Jiwen and Chen, Ken and Asada, Harry and Fu, Chenglong},
  year = {2020},
  month = jul,
  journal = {Journal of Mechanisms and Robotics},
  volume = {12},
  number = {061014},
}

@article{amanhoud2021,
  title = {Contact-Initiated Shared Control Strategies for Four-Arm Supernumerary Manipulation with Foot Interfaces},
  author = {Amanhoud, Walid and Hernandez Sanchez, Jacob and Bouri, Mohamed and Billard, Aude},
  year = {2021},
  month = aug,
  journal = {The International Journal of Robotics Research},
  volume = {40},
  number = {8-9},
  pages = {986--1014},
  publisher = {SAGE Publications Ltd STM},
}

@article{dominijanni2021,
  title = {The Neural Resource Allocation Problem When Enhancing Human Bodies with Extra Robotic Limbs},
  author = {Dominijanni, Giulia and Shokur, Solaiman and Salvietti, Gionata and Buehler, Sarah and Palmerini, Erica and Rossi, Simone and De Vignemont, Frederique and {d'Avella}, Andrea and Makin, Tamar R. and Prattichizzo, Domenico and Micera, Silvestro},
  year = {2021},
  month = oct,
  journal = {Nature Machine Intelligence},
  volume = {3},
  number = {10},
  pages = {850--860},
}

@article{hussain2016,
  title = {The {{Soft-SixthFinger}}: A {{Wearable EMG Controlled Robotic Extra-Finger}} for {{Grasp Compensation}} in {{Chronic Stroke Patients}}},
  shorttitle = {The {{Soft-SixthFinger}}},
  author = {Hussain, Irfan and Salvietti, Gionata and Spagnoletti, Giovanni and Prattichizzo, Domenico},
  year = {2016},
  month = jul,
  journal = {IEEE Robotics and Automation Letters},
  volume = {1},
  number = {2},
  pages = {1000--1006},
}

@article{maimeri2019,
  title = {Design and {{Assessment}} of {{Control Maps}} for {{Multi-Channel sEMG-Driven Prostheses}} and {{Supernumerary Limbs}}},
  author = {Maimeri, Michele and Della Santina, Cosimo and Piazza, Cristina and Rossi, Matteo and Catalano, Manuel G. and Grioli, Giorgio},
  year = {2019},
  month = may,
  journal = {Frontiers in Neurorobotics},
  volume = {13},
  publisher = {Frontiers}
}

@article{tang2022,
  title = {Wearable {{Supernumerary Robotic Limb System Using}} a {{Hybrid Control Approach Based}} on {{Motor Imagery}} and {{Object Detection}}},
  author = {Tang, Zhichuan and Zhang, Lingtao and Chen, Xin and Ying, Jichen and Wang, Xinyang and Wang, Hang},
  year = {2022},
  journal = {IEEE Transactions on Neural Systems and Rehabilitation Engineering},
  volume = {30},
  pages = {1298--1309},
}

@inproceedings{cunningham2018,
  title = {The {{Supernumerary Robotic}} 3rd {{Thumb}} for {{Skilled Music Tasks}}},
  booktitle = {{{IEEE International Conference}} on {{Biomedical Robotics}} and {{Biomechatronics}} ({{Biorob}})},
  author = {Cunningham, James and Hapsari, Anita and Guilleminot, Pierre and Shafti, Ali and Faisal, A. Aldo},
  year = {2018},
  month = aug,
  pages = {665--670},
}

@article{hussain2017,
  title = {Toward Wearable Supernumerary Robotic Fingers to Compensate Missing Grasping Abilities in Hemiparetic Upper Limb},
  author = {Hussain, Irfan and Spagnoletti, Giovanni and Salvietti, Gionata and Prattichizzo, Domenico},
  year = {2017},
  month = dec,
  journal = {The International Journal of Robotics Research},
  volume = {36},
  number = {13-14},
  pages = {1414--1436}
}

@inproceedings{saraiji2018_metarms,
  title = {{{MetaArms}}: {{Body Remapping Using Feet-Controlled Artificial Arms}}},
  shorttitle = {{{MetaArms}}},
  booktitle = {{{Annual ACM Symposium}} on {{User Interface Software}} and {{Technology}}},
  author = {Saraiji, Mhd Yamen and Sasaki, Tomoya and Kunze, Kai and Minamizawa, Kouta and Inami, Masahiko},
  year = {2018},
  month = oct,
  pages = {65--74},
}

@inproceedings{sasaki2017_metalimbs,
  title = {{{MetaLimbs}}: Multiple Arms Interaction Metamorphism},
  shorttitle = {{{MetaLimbs}}},
  booktitle = {{{ACM SIGGRAPH}} 2017 {{Emerging Technologies}}},
  author = {Sasaki, Tomoya and Saraiji, Mhd Yamen and Fernando, Charith Lasantha and Minamizawa, Kouta and Inami, Masahiko},
  year = {2017},
  month = jul,
  pages = {1--2},
}

@article{jiang2025,
  title = {{{CPG-Based Manipulation With Multi-Module Origami Robot Surface}}},
  author = {Jiang, Yuhao and Asmar, Serge El and Wang, Ziqiao and Demirtas, Serhat and Paik, Jamie},
  year = {2025},
  month = may,
  journal = {IEEE Robotics and Automation Letters},
  volume = {10},
  number = {5},
  pages = {4786--4793},
}

@article{mintchev2019,
  title = {A Portable Three-Degrees-of-Freedom Force Feedback Origami Robot for Human--Robot Interactions},
  author = {Mintchev, Stefano and Salerno, Marco and Cherpillod, Alexandre and Scaduto, Simone and Paik, Jamie},
  year = {2019},
  month = dec,
  journal = {Nature Machine Intelligence},
  volume = {1},
  number = {12},
  pages = {584--593},
}

@article{wang2025,
  title = {Surface-Based Manipulation with Modular Foldable Robots},
  author = {Wang, Ziqiao and Demirtas, Serhat and Zuliani, Fabio and Paik, Jamie},
  year = 2026,
  month = jan,
  journal = {npj Robotics},
  volume = {4},
  number = {1},
  pages = {3},
}

@article{luo2025,
  title = {Enhancing {{Human}}--{{Robot Collaboration}}: {{Supernumerary Robotic Limbs}} for {{Object Balance}}},
  shorttitle = {Enhancing {{Human}}--{{Robot Collaboration}}},
  author = {Luo, Jing and Liu, Shiyang and Si, Weiyong and Zeng, Chao},
  year = {2025},
  month = feb,
  journal = {IEEE Transactions on Systems, Man, and Cybernetics: Systems},
  volume = {55},
  number = {2},
  pages = {1334--1347},
}

@inproceedings{llorensbonilla2012,
  title = {Demonstration-Based Control of Supernumerary Robotic Limbs},
  booktitle = {{{IEEE}}/{{RSJ International Conference}} on {{Intelligent Robots}} and {{Systems}}},
  author = {{Llorens - Bonilla}, Baldin and Parietti, Federico and Asada, H. Harry},
  year = {2012},
  month = oct,
  pages = {3936--3942},
}

@book{abdi2016,
  title = {Third Arm Manipulation for Surgical Applications: {{An}} Experimental Study},
  bootktitle = {New Trends in Medical and Service Robots},
  author = {Abdi, E. and Bouri, M. and Himidan, S. and Burdet, E. and Bleuler, H.},
  year = {2016},
  pages = {153--163},
}

@article{guggenheim2020,
  title = {Leveraging the {{Human Operator}} in the {{Design}} and {{Control}} of {{Supernumerary Robotic Limbs}}},
  author = {Guggenheim, Jacob and Hoffman, Rachel and Song, Hanjun and Asada, H. Harry},
  year = {2020},
  month = apr,
  journal = {IEEE Robotics and Automation Letters},
  volume = {5},
  number = {2},
  pages = {2177--2184},
}

@article{penaloza2018,
  title = {{{BMI}} Control of a Third Arm for Multitasking},
  author = {Penaloza, Christian I. and Nishio, Shuichi},
  year = {2018},
  month = jul,
  journal = {Science Robotics},
  volume = {3},
  number = {20},
  pages = {eaat1228}
}

@article{salerno2020,
  title = {Ori-{{Pixel}}, a {{Multi-DoFs Origami Pixel}} for {{Modular Reconfigurable Surfaces}}},
  author = {Salerno, Marco and Paik, Jamie and Mintchev, Stefano},
  year = {2020},
  month = oct,
  journal = {IEEE Robotics and Automation Letters},
  volume = {5},
  number = {4},
  pages = {6988--6995},
}

@article{mete2021,
  title = {Closed-{{Loop Position Control}} of a {{Self-Sensing}} 3-{{DoF Origami Module With Pneumatic Actuators}}},
  author = {Mete, Mustafa and Paik, Jamie},
  year = {2021},
  month = oct,
  journal = {IEEE Robotics and Automation Letters},
  volume = {6},
  number = {4},
  pages = {8213--8220},
}

@misc{churchill1978,
  title = {Anthropometric Source Book.  {{Volume}} 1:  {{Anthropometry}} for Designers},
  shorttitle = {Anthropometric Source Book.  {{Volume}} 1},
  author = {Churchill, E. and Laubach, L. L. and Mcconville, J. T. and Tebbetts, I.},
  year = {1978},
  month = jul,
  urldate = {2025-07-03},
}

@article{namdari2012,
  title = {Defining Functional Shoulder Range of Motion for Activities of Daily Living},
  author = {Namdari, Surena and Yagnik, Gautam and Ebaugh, D. David and Nagda, Sameer and Ramsey, Matthew L. and Williams, Gerald R. and Mehta, Samir},
  year = {2012},
  month = sep,
  journal = {Journal of Shoulder and Elbow Surgery},
  volume = {21},
  number = {9},
  pages = {1177--1183},
}

@article{sardelli2011,
  title = {Functional {{Elbow Range}} of {{Motion}} for {{Contemporary Tasks}}},
  author = {Sardelli, Matthew and Tashjian, Robert Z. and MacWilliams, Bruce A.},
  year = {2011},
  month = mar,
  journal = {JBJS},
  volume = {93},
  number = {5},
  pages = {471},
}

@article{kim2025,
  title = {Supernumerary {{Wearable Soft Manipulator With Modularized Origami Structures}}},
  author = {Kim, Jinho and Cha, Youngsu},
  year = {2025},
  journal = {IEEE/ASME Transactions on Mechatronics},
  pages = {1--11}
}

@article{liao2023,
  title = {Collaborative Workspace Design of Supernumerary Robotic Limbs Base on Multi-Objective Optimization},
  author = {Liao, Ziyu and Chen, Bai and Qian, Zheng and Chang, Tianzuo and Bai, Dongming and Liu, Keming and Xu, JiaJun},
  year = {2023},
  month = jun,
  journal = {Journal of the Brazilian Society of Mechanical Sciences and Engineering},
  volume = {45},
  number = {7},
  pages = {354}
}

@inproceedings{yamamura2023,
  title = {Social {{Digital Cyborgs}}: {{The Collaborative Design Process}} of {{JIZAI ARMS}}},
  shorttitle = {Social {{Digital Cyborgs}}},
  booktitle = {{{CHI Conference}} on {{Human Factors}} in {{Computing Systems}}},
  author = {Yamamura, Nahoko and Uriu, Daisuke and Muramatsu, Mitsuru and Kamiyama, Yusuke and Kashino, Zendai and Sakamoto, Shin and Tanaka, Naoki and Tanigawa, Toma and Onishi, Akiyoshi and Yoshida, Shigeo and Yamanaka, Shunji and Inami, Masahiko},
  year = {2023},
  month = apr,
  pages = {1--19},
}

@inproceedings{nabeshima2019,
  title = {Prosthetic {{Tail}}: {{Artificial Anthropomorphic Tail}} for {{Extending Innate Body Functions}}},
  shorttitle = {Prosthetic {{Tail}}},
  booktitle = {{{Augmented Human International Conference}} 2019},
  author = {Nabeshima, Junichi and Saraiji, MHD Yamen and Minamizawa, Kouta},
  year = {2019},
  month = mar,
  pages = {1--4},
}

@inproceedings{nakabayashi2017,
  title = {Development of {{Evaluation Indexes}} for {{Human-Centered Design}} of a {{Wearable Robot Arm}}},
  booktitle = {{{International Conference}} on {{Human Agent Interaction}}},
  author = {Nakabayashi, Koki and Iwasaki, Yukiko and Iwata, Hiroyasu},
  year = 2017,
  month = oct,
  pages = {305--310},
  publisher = {ACM},
}

@article{leal2025neuromuscular,
  title={Neuromuscular learning and control of auricular muscles for human--machine interfaces},
  author={Leal Pinheiro, Daniel and Faber, Jean and Micera, Silvestro and Shokur, Solaiman},
  journal={The International Journal of Robotics Research},
  pages={02783649251390578},
  year={2025},
  publisher={SAGE Publications Sage UK: London, England}
}

@article{ciullo2020,
  title = {A {{Novel Soft Robotic Supernumerary Hand}} for {{Severely Affected Stroke Patients}}},
  author = {Ciullo, Andrea S. and Veerbeek, Janne M. and Temperli, Eveline and Luft, Andreas R. and Tonis, Frederik J. and Haarman, Claudia J. W. and Ajoudani, Arash and Catalano, Manuel G. and Held, Jeremia P. O. and Bicchi, Antonio},
  year = 2020,
  month = may,
  journal = {IEEE Transactions on Neural Systems and Rehabilitation Engineering},
  volume = {28},
  number = {5},
  pages = {1168--1177}
}

@inproceedings{zhang2024,
  title = {Design and {{Control}} of a {{Soft Supernumerary Robotic Limb Based}} on {{Fiber-Reinforced Actuator}}},
  booktitle = {2024 {{IEEE}}/{{RSJ International Conference}} on {{Intelligent Robots}} and {{Systems}} ({{IROS}})},
  author = {Zhang, Tianyi and Xu, Jiajun and Lu, Yonghua and Zhao, Mengcheng and Huang, Kaizhen and Chen, Bai and Hou, Xuyan and Li, Youfu},
  year = 2024,
  month = oct,
  pages = {9167--9174},
}

@article{bolotnikova2025,
  title = {Optimized User-Guided Motion Control of Modular Robots},
  author = {Bolotnikova, Anastasia and Holdcroft, Kevin and Cerbone, Henry and Belke, Christoph and Ijspeert, Auke and Paik, Jamie},
  year = 2025,
  month = sep,
  journal = {Nature Communications},
  volume = {16},
  number = {1},
  pages = {8675},
}

@article{gao2025,
  title = {Wearable Technologies for Assisted Mobility in the Real World},
  author = {Gao, Shuo and Chen, Jianan and Xia, Yunjia and Li, Xuemeng and Ma, Weihao and Yang, Huixin and Li, Jinchen and Zhou, Xinkai and Jia, Tianyu and Xu, Yuchen and Uchitel, Julie and Ta, Dean and Qi, Peng and Ge, Junbo and Guo, Yi and Qin, Yajie and Kang, Inseung and Xu, Wenyao and Li, He and Chang, Jinke and Zuo, Siming and Wang, Shiwei and Luo, Shan and Gionfrida, Letizia and Hu, Chen and Dong, Shuqin and Guo, Yongxin and Yuan, Yixuan and Zhang, Haixia and Chen, Haotian and Pan, Yu and Dai, Chenyun and Ren, Qinyuan and Loureiro, Rui and Carlson, Tom and Chen, Wei and Zhang, Yuanting and Kyriacou, Panicos and Heidari, Hadi and Nazarpour, Kia and Prodromakis, Themis and Casson, Alexander and Makin, Tamar R. and Cauwenberghs, Gert and Farina, Dario and Zhao, Hubin},
  year = 2025,
  month = dec,
  journal = {Nature Communications},
  volume = {16},
  number = {1},
  pages = {10988},
}

@article{ferroni2025,
  title = {A Multi-Joint Soft Exosuit Improves Shoulder and Elbow Motor Functions in Individuals with Spinal Cord Injury},
  author = {Ferroni, Roberto and D'Avola, Gaetano and Sciarrone, Giorgia and Righi, Gabriele and De Santis, Claudia and Carpaneto, Jacopo and Gandolla, Marta and Del Popolo, Giulio and Micera, Silvestro and Proietti, Tommaso},
  year = 2025,
  month = sep,
  journal = {Nature Machine Intelligence},
  volume = {7},
  number = {9},
  pages = {1390--1402},
}

@article{tricomi2024,
  title = {Soft Robotic Shorts Improve Outdoor Walking Efficiency in Older Adults},
  author = {Tricomi, Enrica and Missiroli, Francesco and Xiloyannis, Michele and Lotti, Nicola and Zhang, Xiaohui and Stefanakis, Marios and Theisen, Maximilian and Bauer, J{\"u}rgen and Becker, Clemens and Masia, Lorenzo},
  year = 2024,
  month = oct,
  journal = {Nature Machine Intelligence},
  volume = {6},
  number = {10},
  pages = {1145--1155},}

\clearpage
\newpage

\setcounter{figure}{0}
\begin{center}
\section*{Supplementary Materials for "Reconfiguration of supernumerary robotic limbs for human augmentation"}

% Author list for the supplement
% Indicate the corresponding authors, but do NOT include institutions here
% It would be nice if the template auto-generated this, but doing so is complicated...

Mustafa Mete$^{1,\dagger}$,
Anastasia Bolotnikova$^{1,2,\dagger}$,
Alexander Schüßler$^{1,\dagger}$
Jamie Paik$^{1,\ast}$ \\
\small{$^{1}$Reconfigurable Robotics Lab, École Polytechnique Fédérale de Lausanne (EPFL), Switzerland}\\
\small{$^{2}$Laboratory for Analysis and Architecture of Systems (LAAS), CNRS, Toulouse, France}\\
\small{$^\ast$Corresponding author. Email:  jamie.paik@epfl.ch.} \\
\small$^\dagger$These authors contributed equally to this work.
\end{center}

% Fill out the numbers for each type of supplementary material,
% and delete any lines that aren't applicable.
% These are just example numbers that don't match the rest of this template.
\subsubsection*{This PDF file includes:}
Figures S1 to S9\\
Tables S1 \\
Captions for Movies S1 to S4\\

\subsubsection*{Other Supplementary Materials for this manuscript:}
Movies S1 to S4

\clearpage

%%%%%%%%%%%%%%%% SUPPLEMENTARY TEXT %%%%%%%%%%%%%%%

%%%%%%%%%%%%%%%% SUPPLEMENTARY FIGURES %%%%%%%%%%%%%%%
\subsection*{Supplementary Figures and Tables}

\begin{figure}[h!]
    \centering
    \includegraphics[width=1\linewidth]{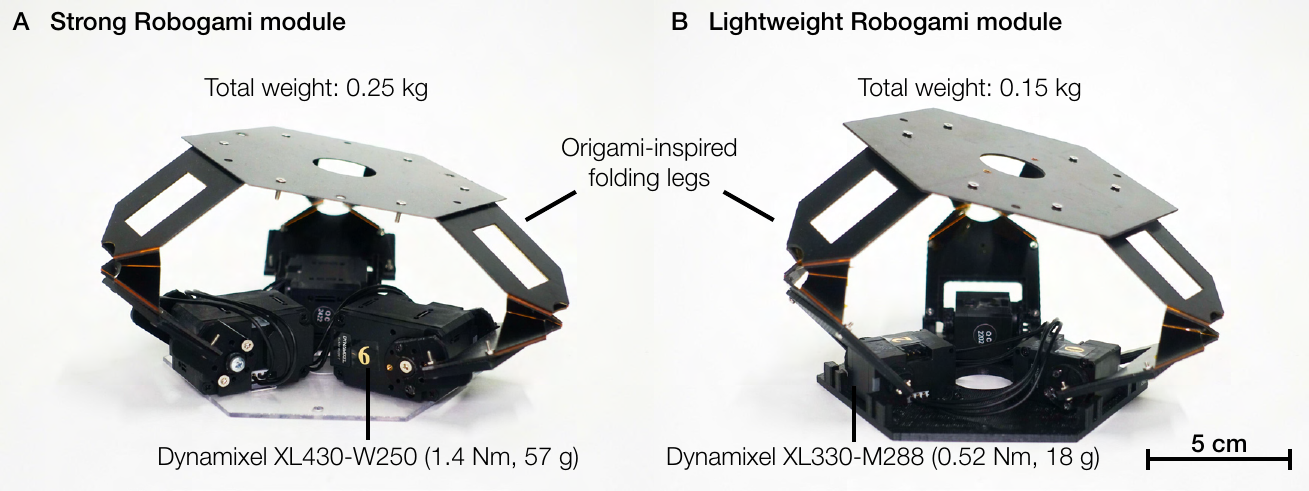}
    \caption{\textbf{Robogami module. }The Robogami module realizes a parallel kinematic structure based on the Canfield joint principle through the use of three origami-inspired folding legs. Each of the legs is connected to a DC motor to realize the actuation of the module. Two versions of the Robogami module are used: (\textbf{A}) A strong but heavier version using a motor with \SI{1.4}{\N\m} torque and a weight of \SI{57}{\gram}. (\textbf{B}) A weaker but lighter version using a motor with \SI{0.52}{\N\m} torque and a weight of \SI{18}{\gram}.}
    \label{fig:supp:robogami_modules}
\end{figure}
\clearpage

\begin{figure}[h!]
    \centering
    \includegraphics[width=1\linewidth]{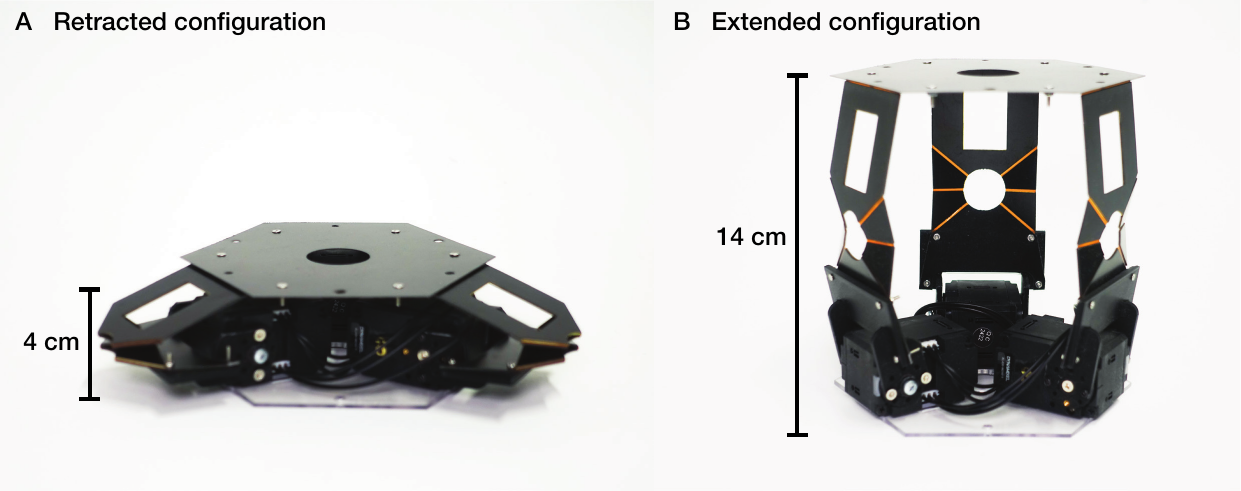}
    \caption{\textbf{Structural reconfiguration of Robogami module. }The parallel kinematic structure of the Robogami module enables reconfiguration from a shape with a \SI{4}{cm} height (\textbf{A}) to a shape with a \SI{14}{cm} height (\textbf{B}). This results in an extension ratio of $3.5$ for a single Robogami module.}
    \label{fig:supp:reconfiguration}
\end{figure}
\clearpage

\begin{figure}[h!]
    \centering
    \includegraphics[width=\linewidth]{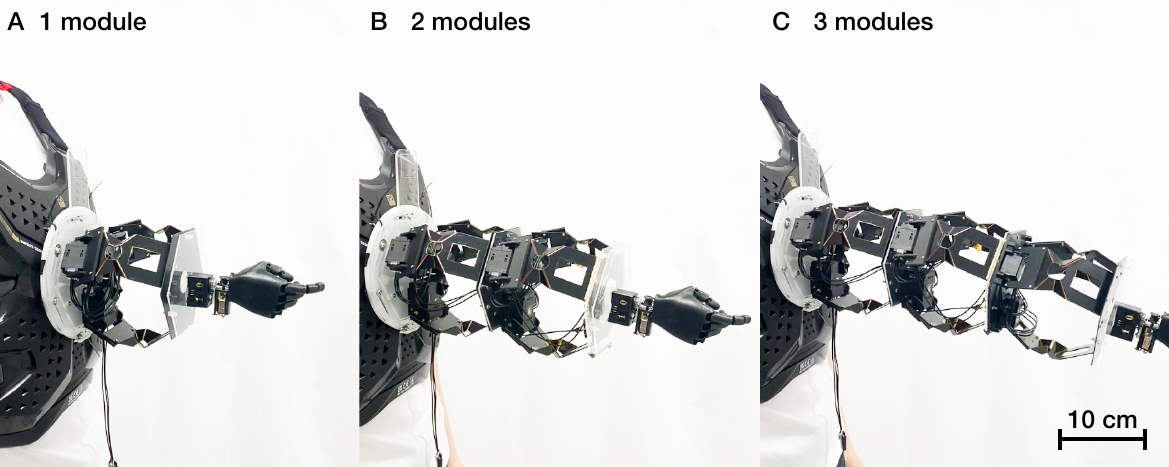}
    \caption{\textbf{Modularity of Robogami Third Arm. }The Robogami modules can be stacked in series to create arms with different configurations (\textbf{A-C}). The modularity is a key feature of the Robogami Third Arm to adapt its workspace and payload capacity to different scenarios and tasks. An underactuated hand is attached to the Robogami Third Arm and can be exchanged for alternative end-effectors depending on the task.}
    \label{fig:supp:modularity}
\end{figure}
\clearpage

\begin{figure}[h!]
    \centering
    \includegraphics[width=\linewidth]{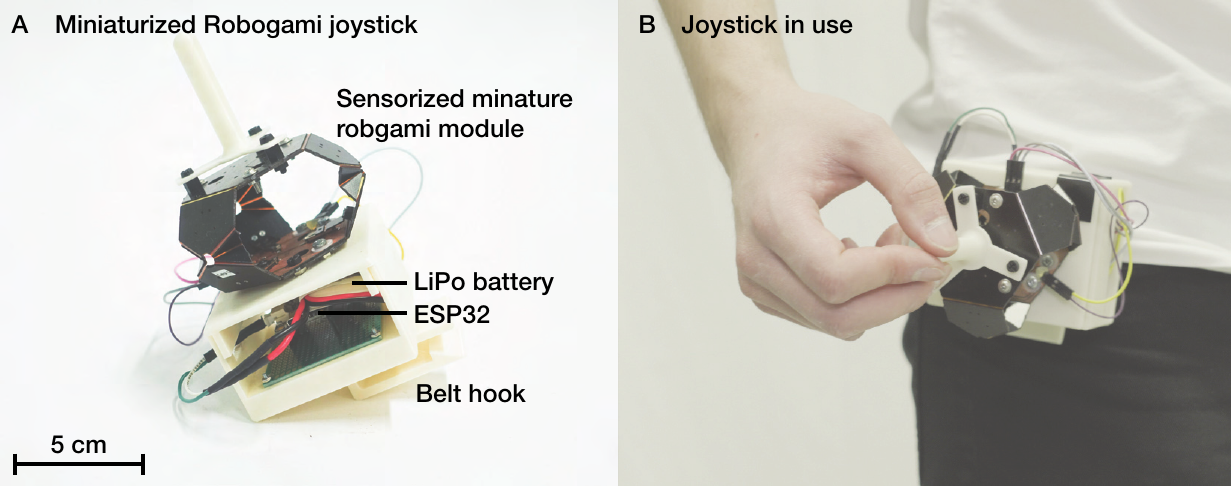}
    \caption{\textbf{Miniaturized Robogami Third Arm joystick. } (\textbf{A}) A 3-DoF joystick based on a sensorized miniaturization of the Robogami Third Arm: The joystick unit includes a lithium-polymer (LiPo) battery for power supply and a microcontroller (ESP32) for Bluetooth Low Energy (BLE) communication with the central controller unit. (\textbf{B}) A belt hook is used to attach the joystick to the belt of the human operator for comfortable use.}
    \label{fig:supp:user_interfaces}
\end{figure}
\clearpage

\begin{figure}[h!]
    \centering
    \includegraphics[width=1\linewidth]{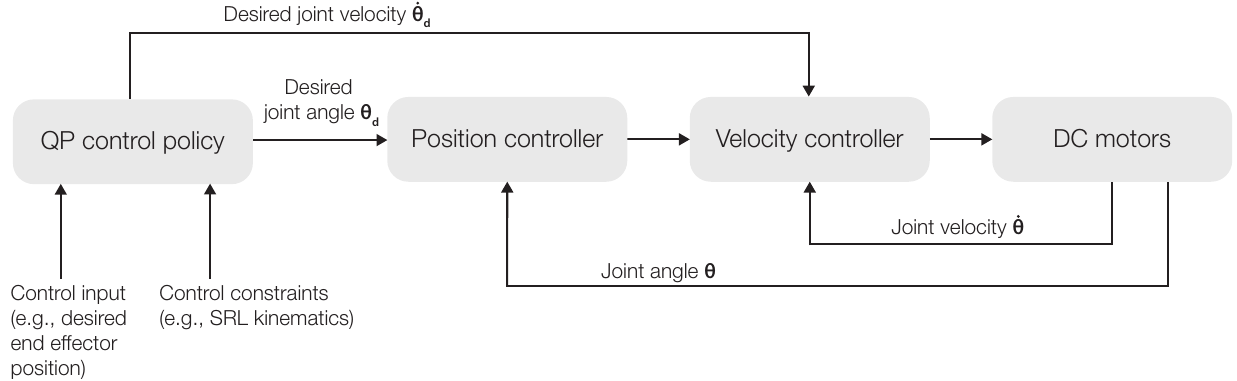}
    \caption{\textbf{Control policy flowchart. }The quadratic programming-based (QP-based) control policy receives control inputs from either an autonomous motion planner or manual control interfaces, together with the control constraints, to compute desired joint-level position and velocity commands. A cascaded PID controller simultaneously regulates joint angles and velocities. Joint angle and velocity limits are applied to the controller outputs to ensure safe execution and prevent damage to the origami-inspired folding legs.}
    \label{fig:supp:low_level_joint_control}
\end{figure}
 \clearpage

\begin{figure}[h!]
    \centering
    \includegraphics[width=1\linewidth]{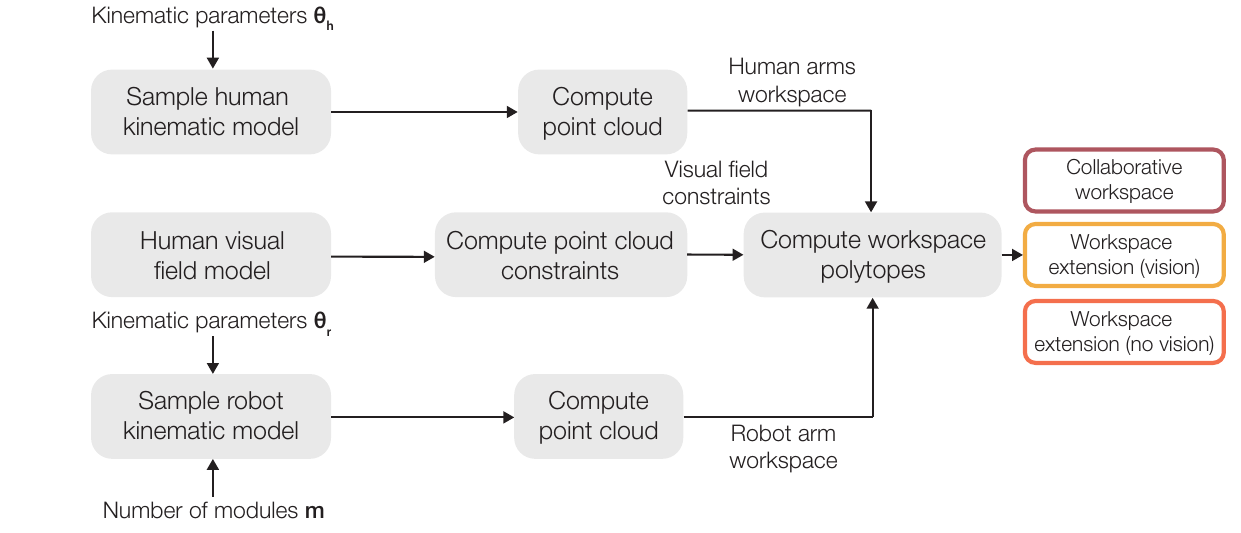}
    \caption{\textbf{Workspace calculation. }The workspaces for supernumerary robotic limb (SRL) configurations are calculated in three steps. In the first step, the human and robot kinematic models are sampled, based on the kinematic parameters and the number of modules of the robot. Based on the samples, point clouds for each human arm and the robot are computed. Moreover, point cloud constraints for the human visual field are computed. Subsequently, convex polytopes for the collaborative, visible extended, and non-visible extended workspaces are computed based on the intersection of the point clouds and constraints.}
    \label{fig:supp:adaptable_configuration}
\end{figure}
 \clearpage

\begin{figure}[h!]
    \centering
    \includegraphics[width=\linewidth]{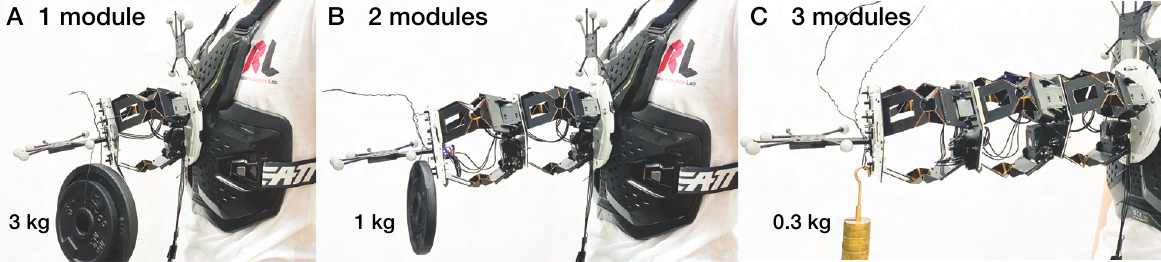}
    \caption{\textbf{Payload capacity. }The payload capacity of reconfigurable SRLs with multiple modules depends on the number of modules $n$, module kinematics $M$, and the module type (strong vs. lightweight module). The payload experiments in a straight-beam configuration evaluate the payload capacity of the Robogami Third Arm with 1, 2, and 3 modules (A, B, and C). The first two modules are strong Robogami modules, while the third module is a lightweight module due to payload limitations. The experiments show a payload capacity of \SI{3}{\kilo\gram} for 1 module, \SI{1}{\kilo\gram} for 2 modules, and \SI{0.3}{\kilo\gram} for 3 modules.}
    \label{fig:supp:payload_capacity}
\end{figure}
 \clearpage

\begin{figure}[h!]
    \centering
    \includegraphics[width=\linewidth]{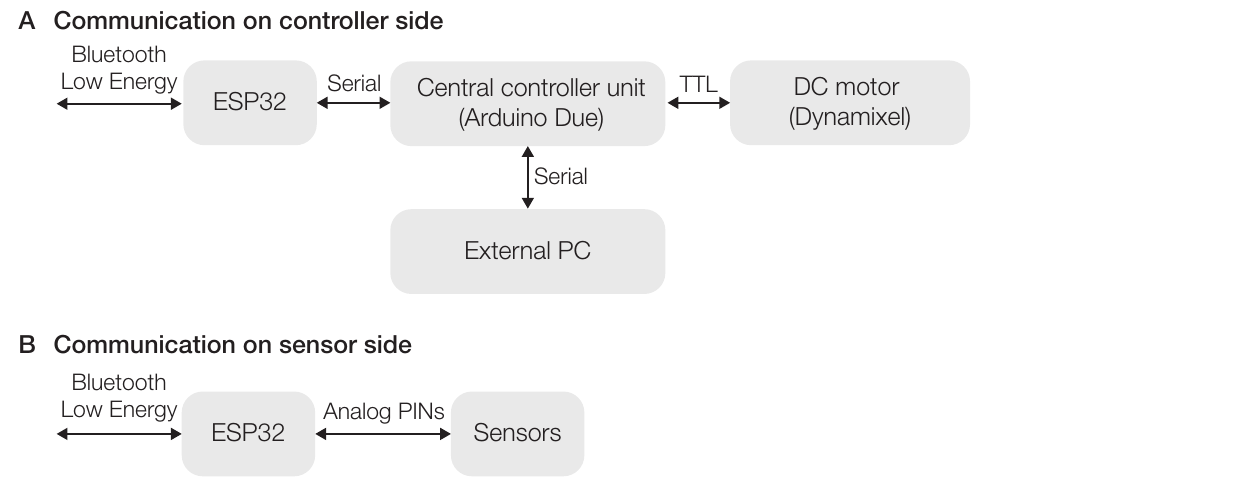}
    \caption{\textbf{Robogami Third Arm communication. }(\textbf{A}) The central controller unit (Arduino Due) acts as the central device that connects all peripherals (DC motors and user interfaces). The joint information from the DC motors (Dynamixel, Robotis) and the joint commands from the low-level joint control on the central controller unit are sent via TTL communication. Using serial communication an external PC for running the high-level quadratic programming-based control policy is connected to the central controller unit. An ESP32 is connected via serial communication to the central controller unit to enable Bluetooth Low Energy (BLE) communication with user interfaces, such as the miniature Robogami joystick. (\textbf{B}) The user interfaces utilize a ESP32 to read the sensor data and send it to the central controller unit via BLE.}
    \label{fig:supp:communication}
\end{figure}
 \clearpage

\begin{figure}[h!]
    \centering
    \includegraphics[width=1\linewidth]{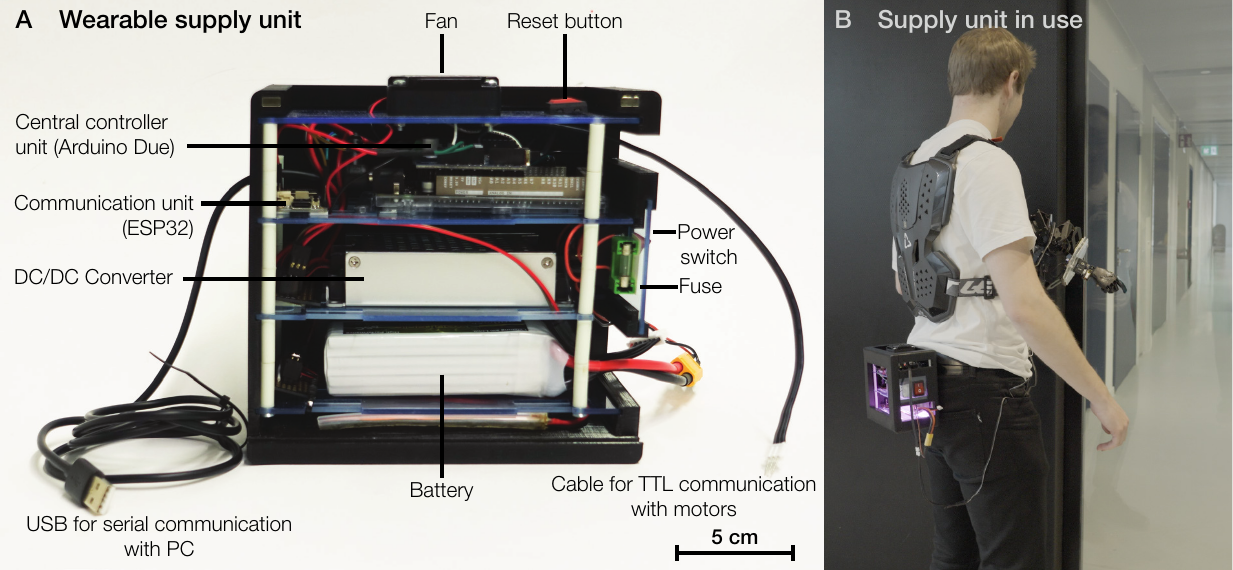}
    \caption{\textbf{Wearable supply unit for untethered operation. }(\textbf{A}) The supply unit includes all necessities for power, communication, and control of the Robogami Third Arm to run it completely untethered on the human body. A lithium-polymer battery with a \SI{1300}{\milli\ampere\hour} capacity supplies the necessary power for running the arm for approximately 3 hours in unloaded scenarios and 1.5 hours in loaded scenarios. A DC/DC converter converts the voltage from the battery to the necessary \SI{12}{\volt} for running the central controller unit (Arduino Due) and the servo motors. A communication unit (ESP32) is connected to the central controller unit for realizing the Bluetooth Low Energy (BLE) connection to the peripheral user interfaces. The supply unit has two cables to connect to external devices: a USB cable for serial communication with an external PC and a cable for TTL communication with the servo motors. (\textbf{B}) The wearable supply unit is attached to the human body for untethered control of the Robogami Third Arm.}
    \label{fig:supp:untethered_box}
\end{figure}

%%%%%%%%%%%%%%%% SUPPLEMENTARY TABLES %%%%%%%%%%%%%%%
\clearpage
\begin{table}[h!]
\caption{Overview of existing supernumerary robotic limbs (SRLs) and the presented Robogami Third Arm with different number of modules.}
\begin{tabular}{lccccc}
\\
\hline
Robot arm & \multicolumn{1}{l}{\begin{tabular}[c]{@{}l@{}} Weight \\ (\SI{}{\gram})\end{tabular}} & \multicolumn{1}{l}{\begin{tabular}[c]{@{}l@{}}Payload \\ capacity \\ (\SI{}{\gram})\end{tabular}} & \multicolumn{1}{l}{\begin{tabular}[c]{@{}l@{}}Load-to-weight \\ ratio (-)\end{tabular}} & \multicolumn{1}{l}{\begin{tabular}[c]{@{}l@{}}Max. arm \\ extension \\ (\SI{}{\milli\meter})\end{tabular}} & \multicolumn{1}{l}{\begin{tabular}[c]{@{}l@{}}Extension \\ ratio (-)\end{tabular}} \\ \hline
Robogami Third Arm  & 250 & 3000 & 12   & 140 & 3.5  \\
(1 strong module) & & & & & \\
Robogami Third Arm  & 500 & 1000 & 2  & 280  & 3.5   \\
(2 strong modules) & & & & & \\
Robogami Third Arm & 650 & 300 & 0.46 & 420  & 3.5 \\
(2 strong + 1 lightweight module) & & & & & \\
Soft Poly-Limb \cite{Nguyen2019_1, Nguyen2019_2} & 1100 & 1500  & 1.36 & 695  & 2 \\
Origami-inspired SRL \cite{Kusunoki2023} & 78 & 200 & 3.56 & 280 & 5.44 \\
Universal Robots UR5e \cite{Ballesteros2023} & 20600 & 5000 & 0.24 & 850 & 1 \\ % (UR5e)
RSRAs \cite{zhang2022} & 3450 & 5000 & 1.45 & 940 & 1 \\ \hline
\end{tabular}
\label{tab:supp:related_work_comparison}
\end{table}
%%%%%%%%%%% CAPTIONS FOR OTHER SUPPLEMENTARY FILES %%%%%%%%%%

\clearpage % Clear all remaining figures and tables then start a new page

\paragraph{Caption for Movie S1.}
\textbf{Reconfiguration for human augmentation summary.}
Reconfiguration of supernumerary robotic limbs improves adaptability to dynamic and unstructured environments. The reconfiguration is highlighted by the Robogami Third Arm based on origami-inspired modules. Using the reconfiguration strategy, the Robogami Third Arm is able to reconfigure its structure (number of modules), as well as its placement on the human body. The control strategy adapts the level of autonomy depending on the task and scenario from manual control to autonomous control. Based on the reconfiguration and the control strategy, the Robogami Third Arm solves diverse tasks of everyday life.

\paragraph{Caption for Movie S2.}
\textbf{Robogami Third Arm overview.}
The Robogami Third Arm consists of origami-inspired robotic modules. This design enables lightweight construction and reconfigurability, supporting compact storage and everyday wearability. The modular design is leveraged for spatial reconfiguration of the arm. The control strategy adapts to varying SRL configurations, scenarios, and tasks.

\paragraph{Caption for Movie S3.}
\textbf{Robogami Third Arm applications.}
The reconfiguration of supernumerary robotic limbs, such as the Robogami Third Arm, solves diverse tasks of everyday life. The Robogami Third Arm stabilizes a cup to free the human hands for other manipulation tasks. It picks up objects from the ground and passes them to the human, so that the human does not need to bend down. Moreover, the Robogami Third Arm assists humans during cooking tasks. While the human prepares the bread, the Robogami Third Arm makes sure that the fondue does not burn by stirring the fondue autonomously on the human's back.

\paragraph{Caption for Movie S4.}
\textbf{Object stabilization experiment.}
The object stabilization highlights the closed-loop control performance of the QP-based control policy (Fig. \ref{fig:5:cup_stabilization_results}). The control policy keeps the end-effector stable, while the human wearer applies motion disturbances to the Robogami Third Arm.

\end{document}